\pdfoutput=1

\documentclass[11pt]{article}

\usepackage[]{acl}

\usepackage{times}
\usepackage{latexsym}

\usepackage[T1]{fontenc}

\usepackage[utf8]{inputenc}

\usepackage{microtype}

\usepackage{latexsym}
\usepackage{url}
\usepackage{dsfont}
\usepackage{acronym}
\usepackage[noend]{algorithmic}
\usepackage{algorithm}
\usepackage{bbm}
\usepackage{beramono}
\usepackage{bm}
\usepackage{booktabs}
\usepackage[skip=0pt,belowskip=0pt,aboveskip=0pt]{caption}
\usepackage[T1]{fontenc}
\usepackage{graphicx}
\usepackage{hyperref}
\usepackage{listings}
\usepackage{multirow}
\usepackage{natbib}
\usepackage{subcaption}
\usepackage{tabularx}
\usepackage{enumerate}
\usepackage[inline]{enumitem}
\usepackage{numprint}
\usepackage[normalem]{ulem}
\usepackage{xspace}
\usepackage{graphicx}
\usepackage{xcolor}
\usepackage{colortbl}
\usepackage{amssymb}
\usepackage{amsmath}
\usepackage{cleveref}
\usepackage{multicol}
\usepackage{xtab}

\usepackage{latexsym}
\usepackage{float}

\usepackage{array}
\usepackage{makecell}
\usepackage{graphicx}
\usepackage{enumitem}

\makeatletter
\renewcommand\subsubsection{\@startsection{subsubsection}{3}{\z@}%
  {0.1\baselineskip \@plus -.1\p@ \@minus -.1\p@}%
  {-3.5\p@}%
  {\ACM@NRadjust{\@subsubsecfont\@adddotafter}}}

\renewcommand\paragraph{\@startsection{paragraph}{4}{\parindent}%
  {0.25\baselineskip \@plus 0\p@ \@minus 0\p@}%
  {-3.5\p@}%
  {\ACM@NRadjust{\@parfont\@adddotafter}}}
\makeatother  

\newcommand{\remembered}{Remembered}
\newcommand{\imagined}{Imagined}

\title{What Makes a Good and Useful Summary? \\ Incorporating Users in Automatic Summarization Research}

\author{Maartje ter Hoeve \\
  University of Amsterdam \\
  \href{mailto:m.a.terhoeve@uva.nl}{m.a.terhoeve@uva.nl} \\\And
  Julia Kiseleva \\
  Microsoft Research \\
  \href{mailto:julia.kiseleva@microsoft.com}{julia.kiseleva@microsoft.com} \\\And
  Maarten de Rijke \\
  University of Amsterdam \\
  \href{mailto:m.derijke@uva.nl}{m.derijke@uva.nl} \\
}

\begin{document}
\maketitle
\begin{abstract}

  Automatic text summarization has enjoyed great progress over the years and is used in numerous applications, impacting the lives of many. Despite this development, there is little research that meaningfully investigates how the current research focus in automatic summarization aligns with users' needs. To bridge this gap, we propose a survey methodology that can be used to investigate the needs of users of automatically generated summaries. Importantly, these needs are dependent on the target group. 
  Hence, we design our survey in such a way that it can be easily adjusted to investigate different user groups. In this work we focus on university students, who make extensive use of summaries during their studies. We find that the current research directions of the automatic summarization community do not fully align with students' needs. Motivated by our findings, we present ways to mitigate this mismatch in future research on automatic summarization: we propose research directions that impact the design, the development and the evaluation of automatically generated summaries.

\end{abstract}


\section{Introduction}
\label{sec:introduction}

The field of automatic text summarization has experienced great progress over the last years, especially since the rise of neural sequence to sequence models~\cite[e.g.,][]{cheng-lapata-2016-neural, see-etal-2017-get, vaswani2017attention}. The introduction of self-supervised transformer language models like BERT~\cite{devlin-etal-2019-bert} has given the field an additional boost~\cite[e.g.,][]{liu2018generating, liu-lapata-2019-text, lewis-etal-2020-bart, xu-etal-2020-discourse}.

The---often \textit{implicit}---goal of automatic text summarization is to generate a condensed textual version of the input document(s), whilst preserving the main message. This is reflected in today's most common evaluation metrics for the task; 
they focus on aspects such as informativeness, fluency, succinctness and 
factuality~\cite[e.g.,][]{lin-2004-rouge, nenkova-passonneau-2004-evaluating, paulus2017deep, narayan-etal-2018-ranking, goodrich2019assessing, wang-etal-2020-asking, xie2021factual}.
The \textit{needs} of the users of the summaries are often not explicitly addressed, despite their importance in \textit{explicit} definitions of the goal of automatic summarization~\cite{jones1999automatic, mani2001automatic}. 
~\citeauthor{mani2001automatic} defines this goal as: ``\emph{to take an information source, extract content from it, and present the most important content to the user in a condensed form and in a manner sensitive to the user's or application's needs.}''

\begin{figure}[tp]
  \begin{subfigure}{\columnwidth}
      \centering
      \includegraphics[width=\columnwidth]{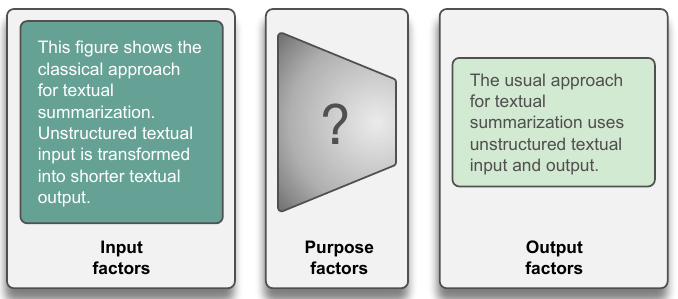}
      \subcaption{Most current automatic text summarization techniques. Left: input document. Right: summary.}
      \label{fig:current_summarization}
  \end{subfigure}

  \begin{subfigure}{\columnwidth}
      \centering
      \includegraphics[width=\columnwidth]{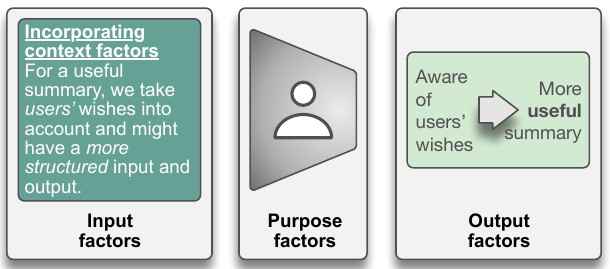}
       \subcaption{Example of summarizing while taking users' wishes and desires into account. Left: input document. Right: summary.}
      \label{fig:useful_summarization}
  \end{subfigure}
\caption{Example of most current summarization techniques vs. summarization while incorporating the users in the process.
}
\label{fig:current_vs_future}
\end{figure}

Different user groups have different needs. Investigating these needs explicitly is critical, given the impact of adequate information transfer~\cite{bennett2012modeling}. We propose a survey methodology to investigate these needs. In designing the survey, we take stock of past work by~\citet{jones1999automatic} who argues that in order to generate useful summaries, one should take the context of a summary into account---a statement that has been echoed by others~\cite[e.g.,][]{mani2001automatic, aries2019automatic}. To do this in a structured manner,~\citeauthor{jones1999automatic} introduces three \textit{context factor} classes: \textit{input factors}, \textit{purpose factors} and \textit{output factors}, which respectively describe the input material, the purpose of the summary, and what the summary should look like. We structure our survey and its implications around these factors. In Figure~\ref{fig:current_vs_future} we give an example of incorporating the context factors in the design of automatic summarization methods.

Our proposed survey can be flexibly adjusted to different user groups. Here we turn our focus to university students as a first stakeholder group.
University students are a particularly relevant group to focus on first, as they benefit from using pre-made summaries in a range of study activities~\cite{reder1980comparison}, but the desired characteristics of these pre-made summaries have not been extensively investigated. We use the word \textit{pre-made} to differentiate such summaries from the ones that users write themselves. Automatically generated summaries fall in the pre-made category, and should thus have the characteristics that users wish for pre-made summaries.

Motivated by our findings, we propose important future research directions that directly impact the design, development, and evaluation of automatically generated summaries. We contribute the following:

\begin{enumerate}[leftmargin=*,label=\textbf{C\arabic*},nosep]
\item We design a survey that can be easily adapted and reused to investigate and understand the needs of the wide variety of users of automatically generated summaries;
\item We develop a thorough understanding of how automatic summarization can optimally benefit users in the educational domain, which leads us to unravel important and currently underexposed research directions for automatic summarization;
\item We propose a new, feasible and comprehensive evaluation methodology to explicitly evaluate the usefulness of a generated summary for its intended purpose.
\end{enumerate}


\section{Related work}
\label{sec:related_work}

In Section~\ref{sec:introduction} we introduced the context factors as proposed by~\citet{jones1999automatic}. Each context factor class can be divided into more fine-grained subclasses. To ensure the flow of the paper, we list an overview in Appendix~\ref{sec:appendix_overview_summary_factors}. Below, we explain and use the context factors and their fine-grained subclasses to structure the related work. 
As our findings have implications for the evaluation of automatic summarization, we also discuss evaluation methods.
Lastly, we discuss the use-cases of automatic summaries in the educational domain.

\subsection{Automatic text summarization}
\label{sec:automatic_summarization}

\textbf{Input factors.}
We start with the fine-grained input factor \textit{unit}, which describes how many sources are to be summarized at once, and the factor \textit{scale}, which describes the length of the input data. 
These factors are related to the difference between single and multi-document summarization~\cite[e.g.,][]{chopra-etal-2016-abstractive, cheng-lapata-2016-neural, wang2016sentence, yasunaga-etal-2017-graph, nallapati2017summarunner, narayan-etal-2018-ranking, liu-lapata-2019-text}. \textit{Scale} plays an important role when material shorter than a single document is summarized, such as sentence summarization~\cite[e.g.,][]{rush-etal-2015-neural}. Regarding the \textit{genre} of the input material, most current work focuses on the news domain or Wikipedia~\citep[e.g.,][]{sandhaus2008new, hermann2015teaching, koupaee2018wikihow, liu2018generating, narayan2018don}. A smaller body of work addresses different input genres, such as scientific articles~\cite[e.g.,][]{cohan-etal-2018-discourse}, forum data~\cite[e.g.,][]{volske-etal-2017-tl}, opinions~\cite[e.g.,][]{amplayo-lapata-2020-unsupervised} or dialogues~\cite[e.g.,][]{liu-etal-2021-topic-aware}. These differences are also closely related to the input factor \textit{subject type}, which describes the difficulty level of the input material. The factor \textit{medium} refers to the input language. Most automatic summarization work is concerned with English as language input, although there are exceptions, such as Chinese~\cite[e.g.,][]{hu-etal-2015-lcsts} or multilingual input~\cite{ladhak-etal-2020-wikilingua}. The last input factor is \textit{structure}. Especially in recent neural approaches, explicit structure of the input text is often ignored. Exceptions include graph-based approaches, where implicit document structure is used to summarize a document~\cite[e.g.,][]{tan-etal-2017-abstractive, yasunaga-etal-2017-graph}, and summarization of tabular data~\cite[e.g.,][]{zhang2020tabular} or screenplays~\cite[e.g.,][]{papalampidi-etal-2020-screenplay}.

\noindent \textbf{Purpose factors.} Although identified as the most important context factor class by~\citet{jones1999automatic}---and followed by, for example, \citet{mani2001automatic}---purpose factors do not receive a substantial amount of attention.
There are some exceptions, e.g., query-based summarization~\cite[e.g.,][]{nema-etal-2017-diversity, litvak2017query}, question-driven summarization~\cite[e.g.,][]{deng-etal-2020-multi}, personalized summarization~\cite[e.g.,][]{moro2012personalized} and interactive summarization~\cite[e.g.,][]{hirsch-etal-2021-ifacetsum}. They take the \textit{situation} and the \textit{audience} into account. The \textit{use}-cases of the generated summaries are also clearer in these approaches. 

\noindent \textbf{Output factors.}
We start with the output factors \textit{style} and \textit{material}. The latter is concerned with the degree of coverage of the summary. Most generated summaries have an \textit{informative} style and cover most of the input material. There are exceptions, e.g., the XSum dataset~\cite{narayan2018don} which constructs summaries of a single sentence and is therefore more \textit{indicative} in terms of style and inevitably less of the input material is covered. Not many summaries have a \textit{critical} or \textit{aggregative} style. Aggregative summaries put different source texts in relation to each other, to give a topic overview. Most popular summarization techniques focus on a \textit{running} \textit{format}. Work on template-based~\cite[e.g.,][]{cao-etal-2018-retrieve} and faceted~\cite[e.g.,][]{meng-etal-2021-bringing} summarization follows a more \textit{headed} (structured) \textit{format}. \citet{falke2017bringing} build concept maps and~\citet{wu2020extracting} make knowledge graphs.
The difference between abstractive and extractive summarization is likely the best known distinction in output type~\cite[e.g.,][]{nallapati2017summarunner, see-etal-2017-get, narayan-etal-2018-ranking, gehrmann-etal-2018-bottom, liu-lapata-2019-text}, although it is not entirely clear which output factor best describes the difference.

\noindent In Section~\ref{sec:implications} we use the context factors to identify future research directions, based on the difference between our findings and the related work.

\subsection{Evaluation}
\label{sec:evaluation}
Evaluation methods for automatic summarization can be grouped in \textit{intrinsic} vs.\ \textit{extrinsic} methods~\cite{mani2001evaluation}. Intrinsic methods evaluate the model itself, e.g., on informativeness or fluency~\cite{paulus2017deep, liu-lapata-2019-text}. Extrinsic methods target how a summary performs when used for a 
task~\cite{dorr-etal-2005-methodology, wang-etal-2020-asking}. Extrinsic methods are resource intensive, 
explaining the popularity of intrinsic methods.

Evaluation methods can also be grouped in \textit{automatic} vs.\ \textit{human} evaluation methods. Different automatic metrics have been proposed, such as Rouge~\cite{lin-2004-rouge} and BERTScore~\cite{zhang2019bertscore} which respectively evaluate lexical and semantic similarity. Other methods use an automatic question-answering evaluation methodology~\cite{wang-etal-2020-asking, durmus-etal-2020-feqa}. Most human evaluation approaches evaluate intrinsic factors such as informativeness, readability and conciseness~\cite{duc2003, nallapati2017summarunner, paulus2017deep, liu-lapata-2019-text}---factors that are difficult to evaluate automatically. There are also some extrinsic human evaluation methods, where judges are asked to perform a certain task based on the summary~\cite[e.g.,][]{narayan-etal-2018-ranking}.
So far, \textit{usefulness}\footnote{We follow the definition of the English Oxford Learner's Dictionary  ({\scriptsize \url{www.oxfordlearnersdictionaries.com/definition/english/}}) for usefulness: \textit{``the fact of being useful or possible to use''}, where \textit{useful} is defined as \textit{``that can help you to do or achieve what you want''}.} has not been evaluated in a feasible and comprehensive manner, whereas it is an important metric to evaluate whether summaries fulfil users' needs.
Therefore, we bridge the gap by introducing a feasible and comprehensive evaluation methodology to evaluate usefulness.

\subsection{Automatic summarization for education}
\label{sec:related_work_education}

Summaries play a prominent role in education. \citet{reder1980comparison} find that students who use a pre-made summary score better on a range of study activities than students who do not use such a summary. As the quality of automatically generated summaries increases~\cite[e.g.,][]{lewis-etal-2020-bart, xu-etal-2020-discourse}, so does the potential to use them in the educational domain, especially given the increasing importance of digital tools and devices for education~\cite{luckin2012decoding, hashim2018application}. With these developments in mind, it is critical that educators are aware of the pedagogical implications; they need to understand how to best make use of all new possibilities~\cite{hashim2018application, amhag2019teacher}.
The outcomes of our survey result in concrete suggestions for developing methods for automatic summarization in the educational domain, whilst taking students' needs into account.


\section{Survey Procedure and Participants}
\label{sec:method}

Here we detail our survey procedure. For concreteness, we present the details with our intended target group in mind. The context factors form the backbone of our survey and the setup can be easily adjusted to investigate the needs of different target groups. For example, we ask participants about a pre-made summary for a recent study activity, but it is straightforward to adapt this to a different use-case that is more suitable for other user groups.

\subsection{Participants}
\label{sec:participants}

We recruited participants among students at universities across the Netherlands by contacting ongoing courses and student associations, and by advertisements on internal student websites. As incentive, we offered a ten euro shopping voucher to ten randomly selected participants.

\begin{figure}
    \centering
     \begin{subfigure}[b]{\columnwidth}
         \centering
         \includegraphics[width=\columnwidth]{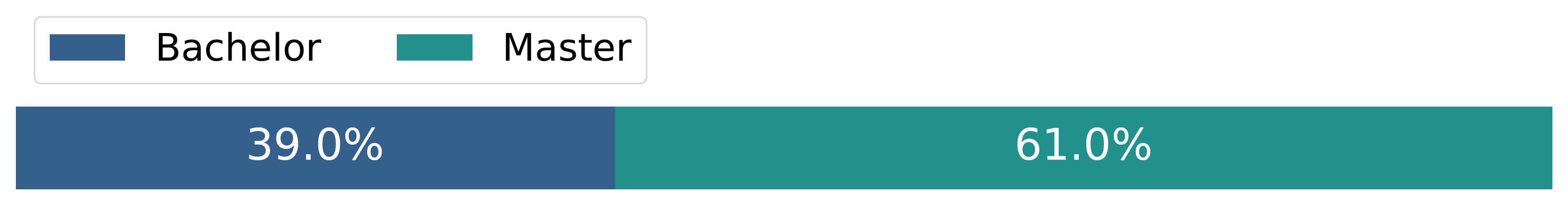}
         \caption{Study levels.}
         \label{fig:study_levels}
     \end{subfigure}
     \begin{subfigure}[b]{\columnwidth}
         \centering
         \includegraphics[width=\columnwidth]{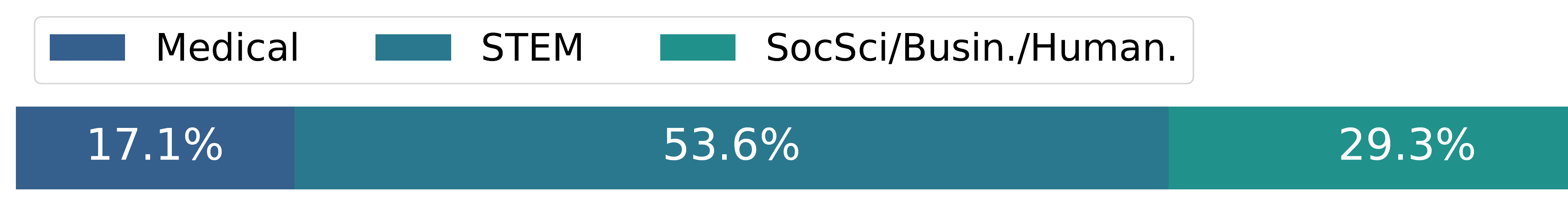}
         \caption{Study backgrounds.}
         \label{fig:study_backgrounds}
     \end{subfigure}
     \caption{Participant details.}
     \label{fig:participants}
  \end{figure}

A total of $118$ participants started the survey and $82$ completed the full survey, resulting in a $69.5\%$ completion rate.
We only include participants who completed the study in our analysis. Participants spent $10$ minutes on average on the survey. In the final part of our survey we ask participants to indicate their current level of education and main field of study. The details are given in Figure~\ref{fig:participants}.

\subsection{Survey procedure}
\label{sec:survey}

Figure~\ref{fig:survey_overview_small} shows a brief overview of our survey procedure. A detailed account is given in Appendix~\ref{sec:appendix_survey_overview}. 
We arrived at the final survey version after a number of pilot runs where we ensured participants understood their task and all questions. We ran the survey with SurveyMonkey ({\small \url{surveymonkey.com}}). A verbatim copy is included in Appendix~\ref{sec:appendix_literal_survey_overview} and released under CC BY license.\footnote{\url{https://github.com/maartjeth/survey_useful_summarization}}
\\

\noindent \textbf{Introduction.}
The survey starts with an introduction where we explain what to expect, how we process the data and that participation is voluntary. After participants agree with this, an explanation of the term \textit{pre-made summary} follows. As we do not want to bias participants by stating that the summary was automatically generated, we explain that the summary can be made by anyone, e.g., a teacher, a good performing fellow student, the authors of the original material, or a computer. Recall that an automatically generated summary is a pre-made summary. Hence, our survey identifies the characteristics an automatically generated summary should have. We also give examples of types of pre-made summaries; based on the pilot experiments we noticed that people missed this information. We explicitly state that these are just examples and that participants can come up with any example of a helpful pre-made summary. 

\noindent \textbf{Context factors.}
In the main part of our survey we focus on the context factors. First, we ask participants whether they have made use of a pre-made summary in one of their recent study activities. If so, we ask them to choose the study activity where a summary was most useful. We call this group the \emph{\remembered{} group}, as they describe an existing summary from memory. If people indicate that they have not used a pre-made summary in one of their recent study activities, we ask them whether they can imagine a situation where a pre-made summary would have been helpful. If not, we ask them why not and lead them to the final background questions and closing page. If yes, we ask them to keep this imaginary situation in mind for the rest of the survey. We call this group the \emph{\imagined{} group}.

Now we ask the \remembered{} and \imagined{} groups about the input, purpose and output factors of the summary they have in mind. We ask questions for each of the context factor subclasses that we discussed in Section~\ref{sec:related_work}.
At this point, the two groups are in different branches of the survey. The difference is mainly linguistically motivated: in the \imagined{} group we use verbs of probability instead of asking to describe an existing situation. Some questions can only be asked in the \remembered{} group, e.g., how helpful the summary was.

In the first context factor question we ask what the study material consisted of. We give a number of options, as well as an `other' checkbox. To avoid position bias, all answer options for multiple choice and multiple response questions in the survey are randomized, with the `other' checkbox always as the last option. If participants do not choose the `mainly text' option, we tell them that we focus on textual input in the current study\footnote{Different modalities are also important to investigate, but we leave this for future work to ensure clarity of our results.} and ask whether they can think of a situation where the input did consist of text. If not, we lead them to the background questions and closing page. If yes, they proceed to the questions that give us a full overview of the input, purpose and output factors of the situation participants have in mind. Finally, we ask the \remembered{} group to suggest how their described summary could be turned into their ideal summary. We then ask both groups for any final remarks about the summary or input material.

\begin{figure}[t]
    \centering
  \includegraphics[width=\columnwidth]{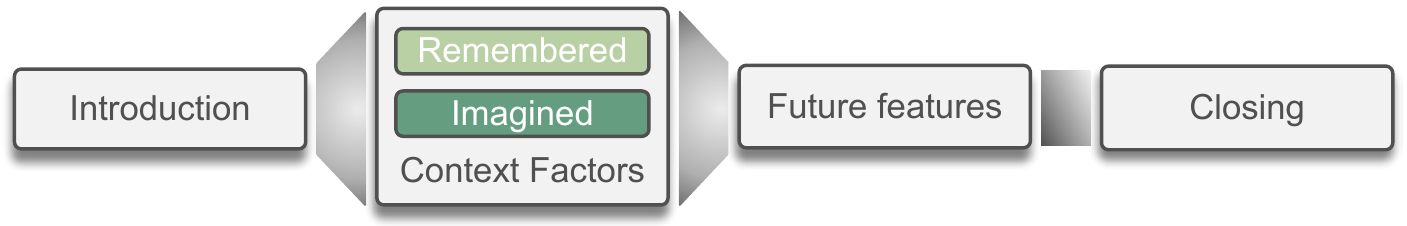}
  \caption{Overview of the survey procedure.}
  \label{fig:survey_overview_small}
  \vspace{-1pt}
  \end{figure}

\noindent \textbf{Trustworthiness and future features questions.} So far we have included the possibility that the summary was machine-generated, but also explicitly included other options so as not to bias participants. At this point we acknowledge that machine-generated summaries could give rise to additional challenges and opportunities. Hence, we include some exploratory questions to get an understanding of the trust users would have in machine-generated summaries and to get ideas for the interpretation of the context factors in
exploratory settings.

For the first questions we tell participants to imagine that the summary was made by a computer, but contained all needs identified in the first part of the survey. We then ask them about trust in computer- and human-generated summaries.
Next, we ask them to imagine that they could interact with the computer program that made the summary in the form of a digital assistant. We tell them not to feel restricted by the capabilities of today's digital assistants.
The verbatim text is given in Appendix~\ref{sec:appendix_literal_survey_overview}.
We ask participants to select the three most and the three least useful features for the digital assistant, similar to~\citet{ter2020conversations}.

\begin{figure*}[t]
  \centering
   \begin{subfigure}[b]{0.31\textwidth}
       \centering
       \caption{\textit{Medium:} The study material consisted of (MC)}
       \vspace{-2pt}
       \includegraphics[clip,trim=0mm 0mm 0mm 10mm,width=\textwidth]{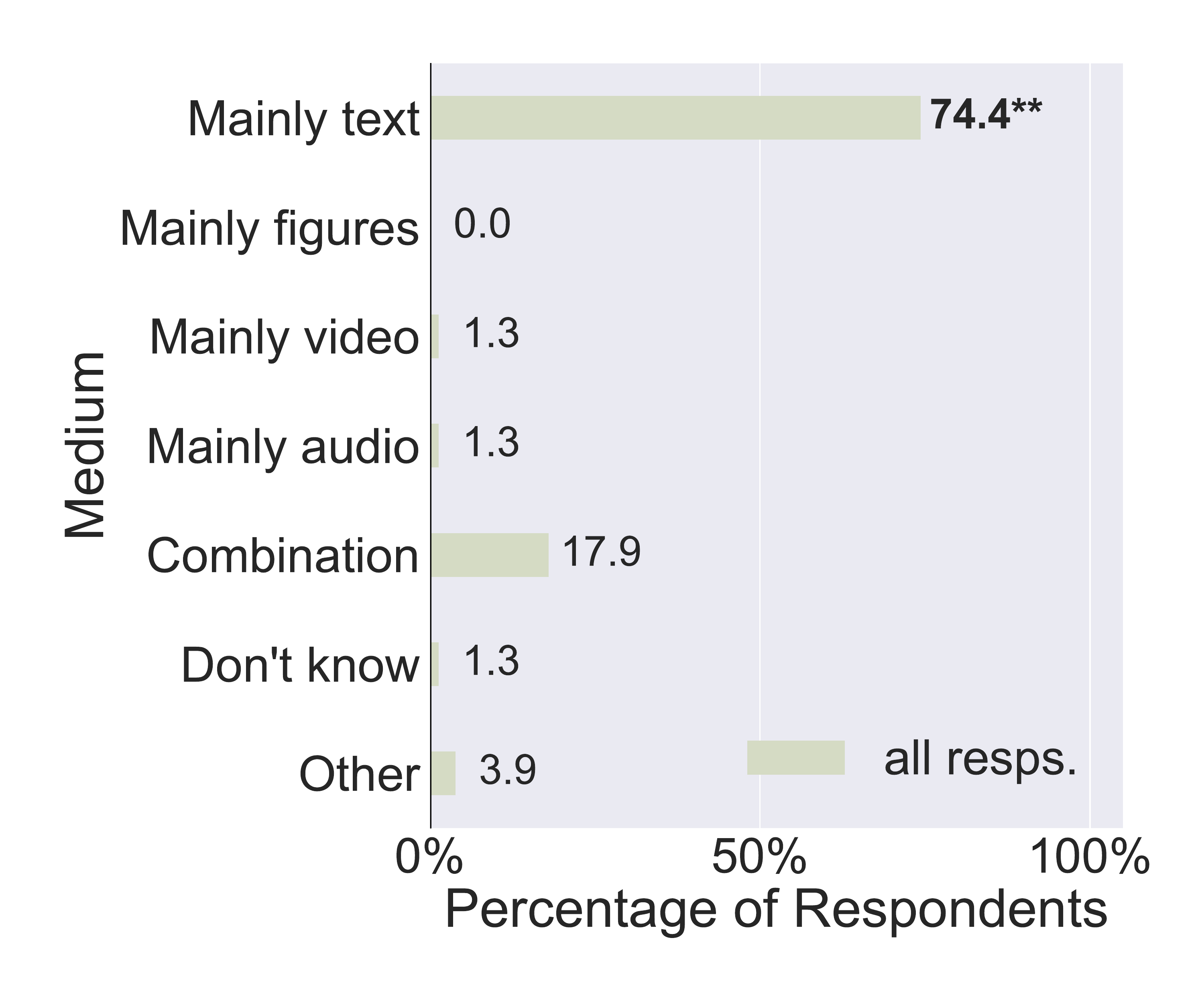}
       \label{fig:input_medium}
   \end{subfigure}\quad
   \begin{subfigure}[b]{0.31\textwidth}
       \centering
       \caption{\textit{Scale / Unit:} What was the length of the study material? (MC)}
       \vspace{-2pt}
       \includegraphics[clip,trim=0mm 0mm 0mm 10mm,width=\textwidth]{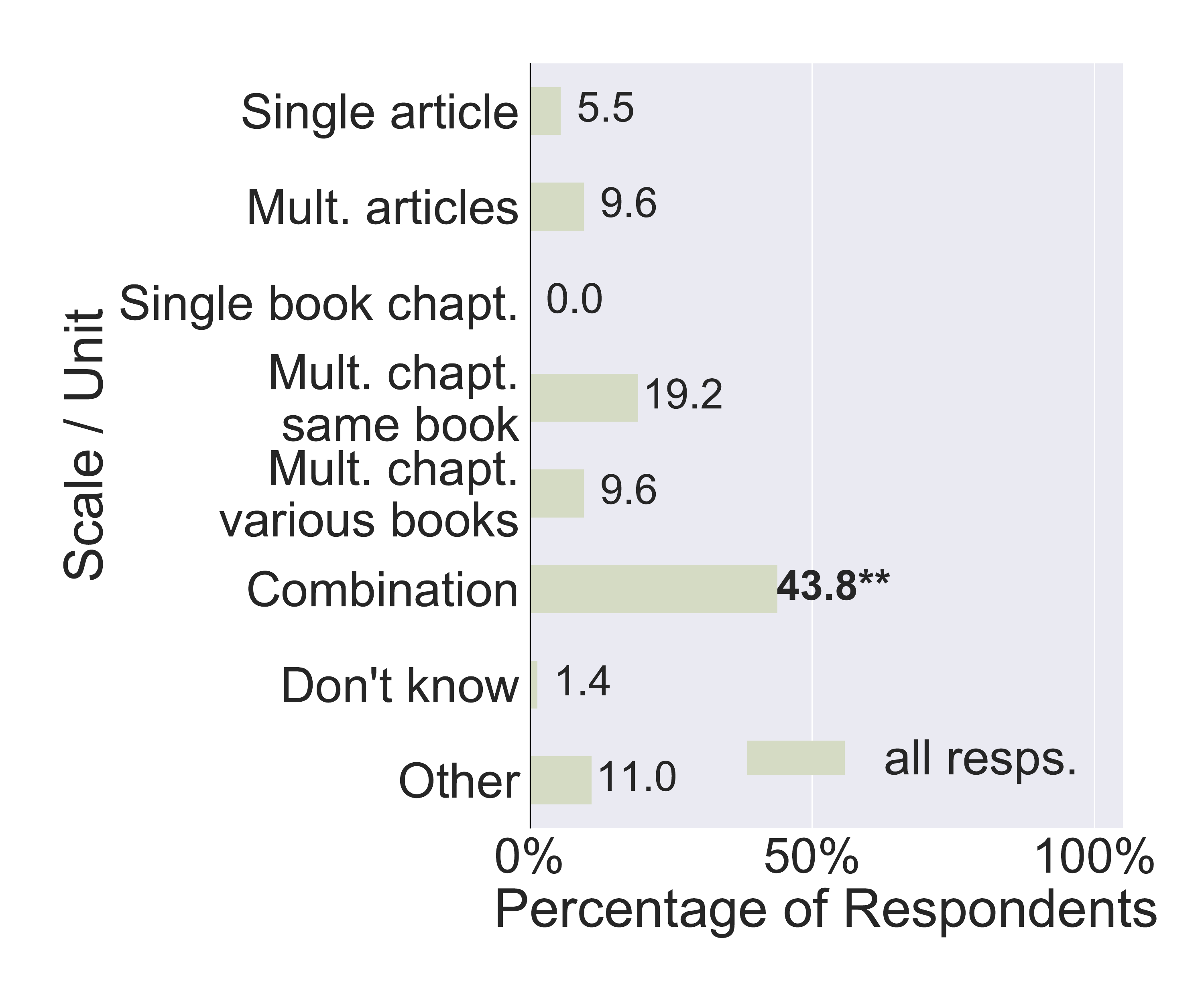}
       \label{fig:input_scale_unit}
   \end{subfigure}\quad
   \begin{subfigure}[b]{0.31\textwidth}
       \centering
       \caption{\textit{Genre:} What was the genre of the study material? (MC)}
       \vspace{-2pt}
       \includegraphics[clip,trim=0mm 0mm 0mm 10mm,width=\textwidth]{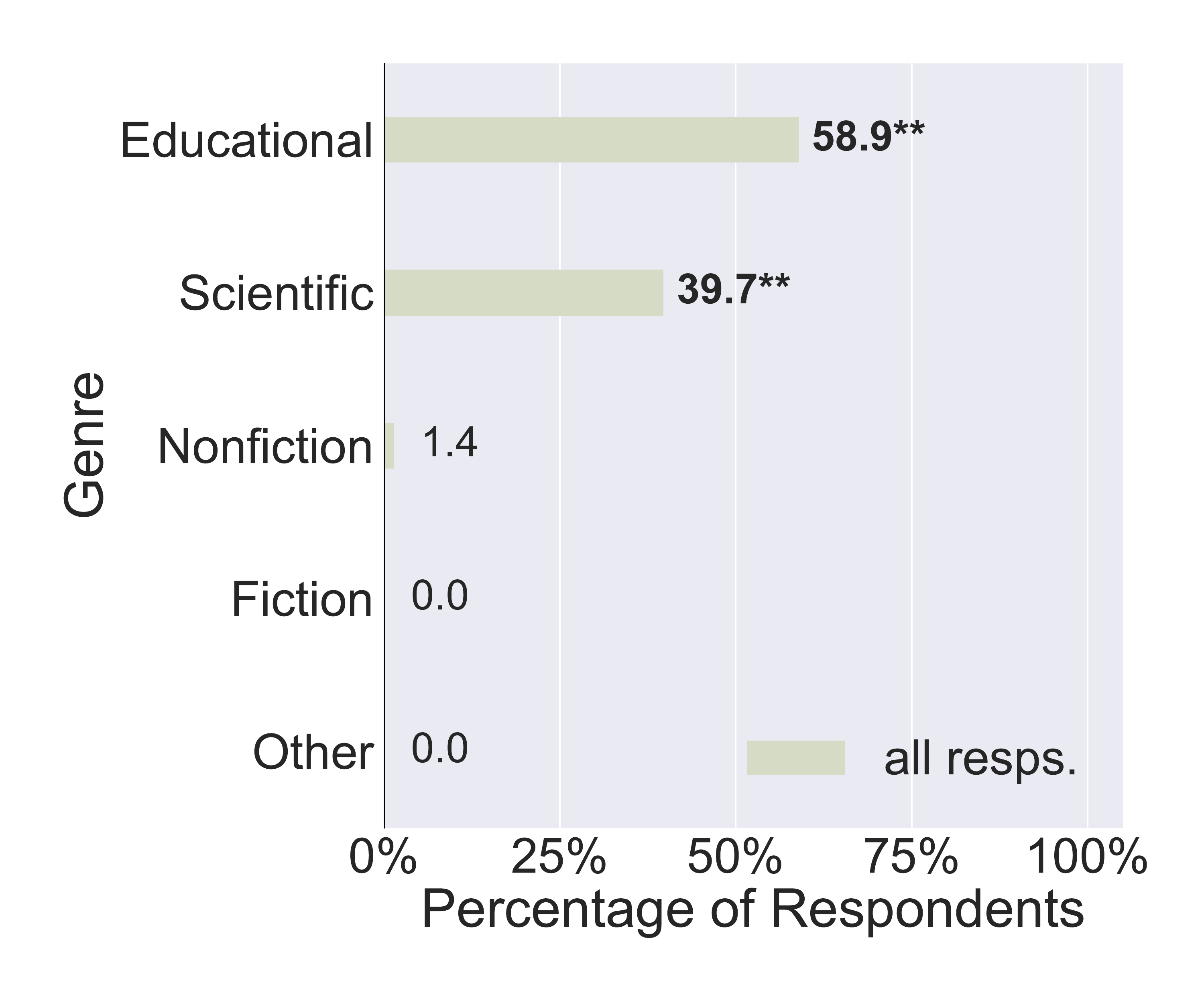}
       \label{fig:input_genre}
   \end{subfigure}

   \vspace*{-1.5em}

   \begin{subfigure}[b]{0.31\textwidth}
       \centering
       \caption{\textit{Subject Type:} How would you classify the difficulty level of the study material? (MC)}
       \vspace{-3pt}
       \includegraphics[clip,trim=0mm 0mm 0mm 10mm,width=\textwidth]{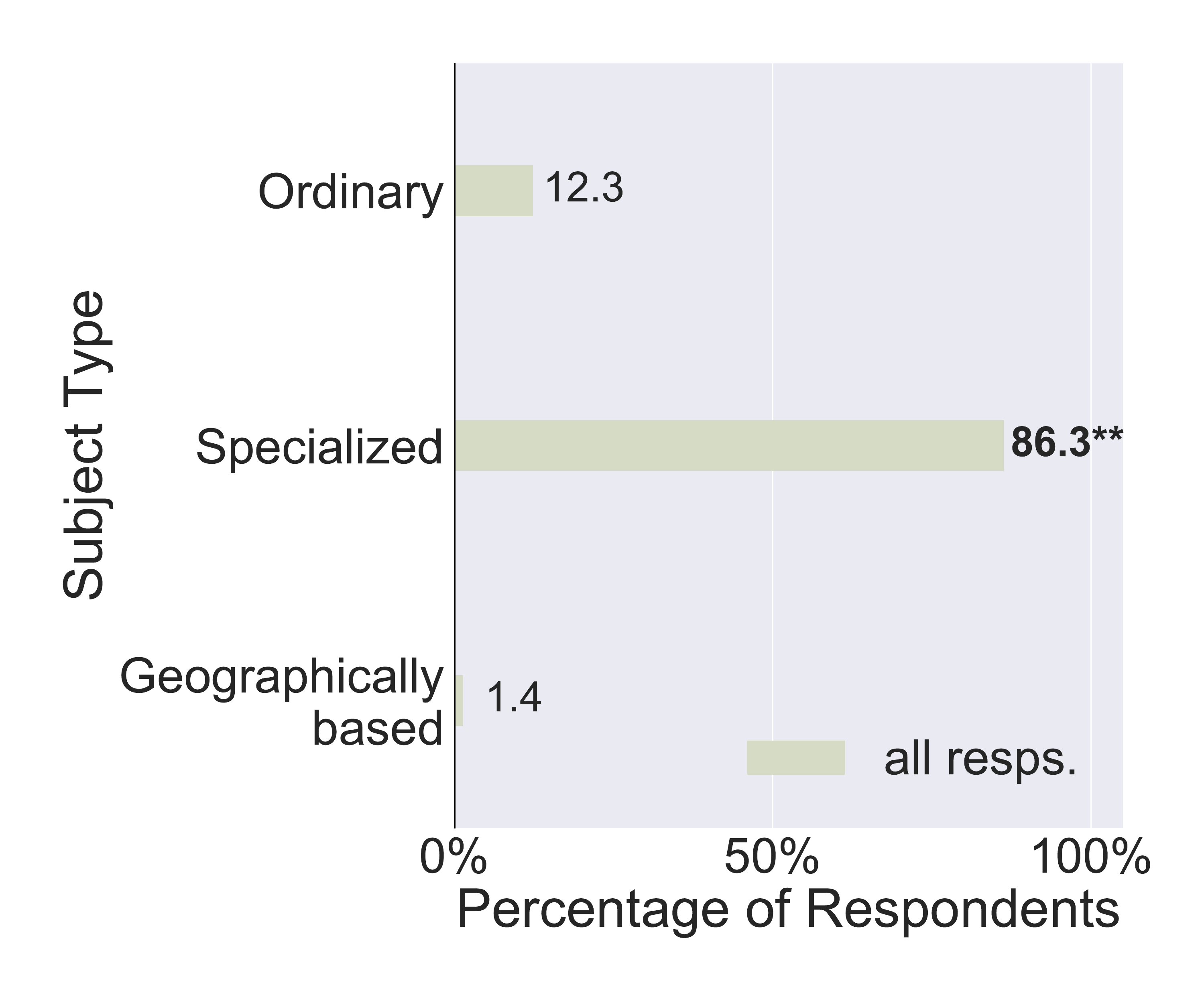}
       \label{fig:input_subject_type}
    \end{subfigure}\quad
    \begin{subfigure}[b]{0.31\textwidth}
        \centering
        \caption{\textit{Structure:} How was the study material structured? (MR) \\}
        \vspace{-3pt}
        \includegraphics[clip,trim=0mm 0mm 0mm 10mm,width=\textwidth]{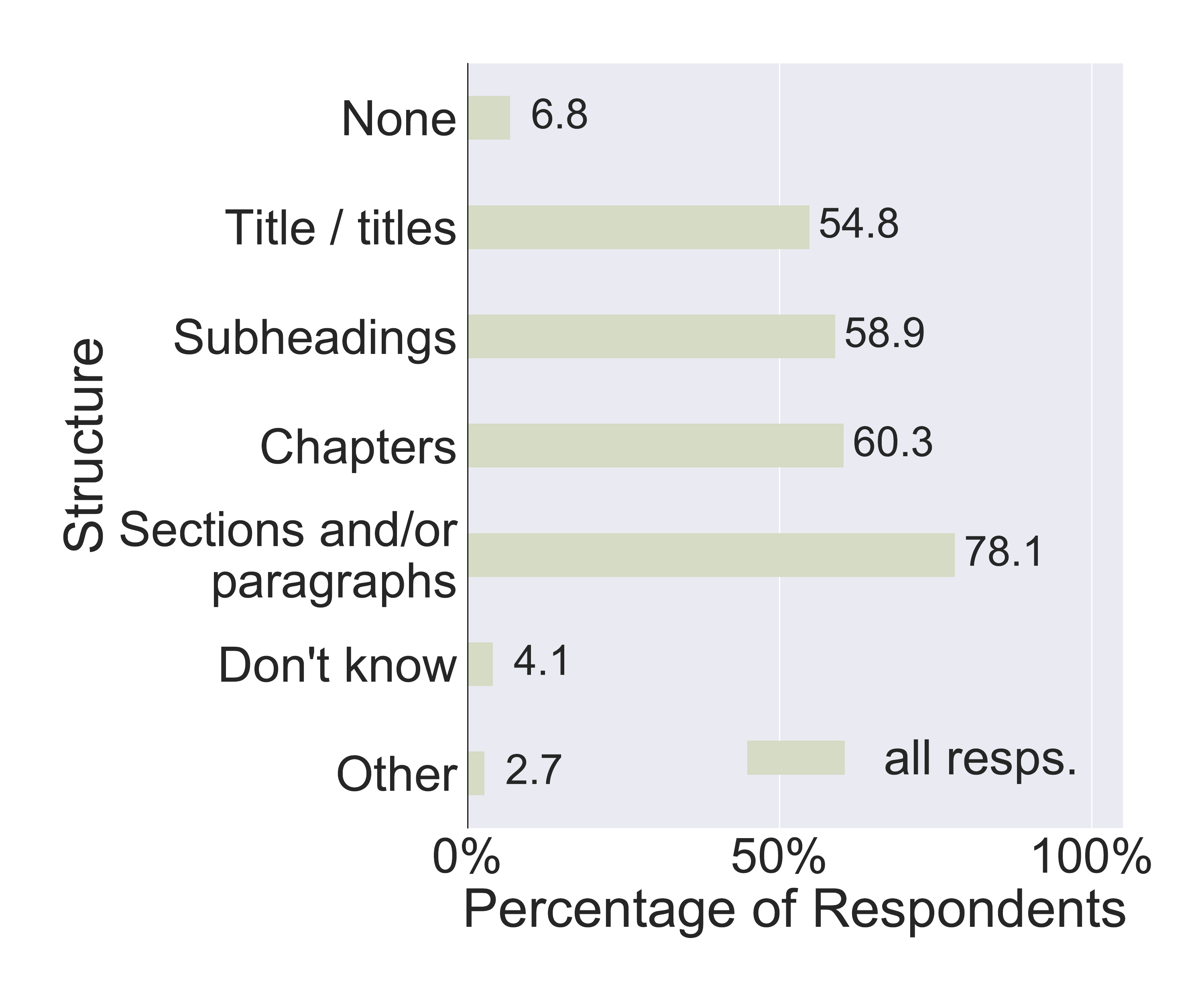}
        \label{fig:input_structure}
    \end{subfigure}
\vspace{-1.5em}
\caption{Results for the \textbf{\emph{input factor}} questions. Specific input factor in italics. Answer type in brackets: MC = Multiple Choice, MR = Multiple Response. \textbf{**} indicates significance ($\chi^2$), after Bonferroni correction, with $p\ll 0.001$. If two options are flagged with \textbf{**}, these options are not significantly different from each other, yet both have been chosen significantly more often than the other options.}
\label{fig:input_factors}
\end{figure*}

\begin{figure*}[!h]
  \begin{subfigure}[b]{0.31\textwidth}
      \centering
      \caption{\textit{Situation (1):} What was the goal of this study activity? (MC) \\ \\}
      \vspace{-2pt}
      \includegraphics[clip,trim=0mm 0mm 0mm 10mm,width=\textwidth]{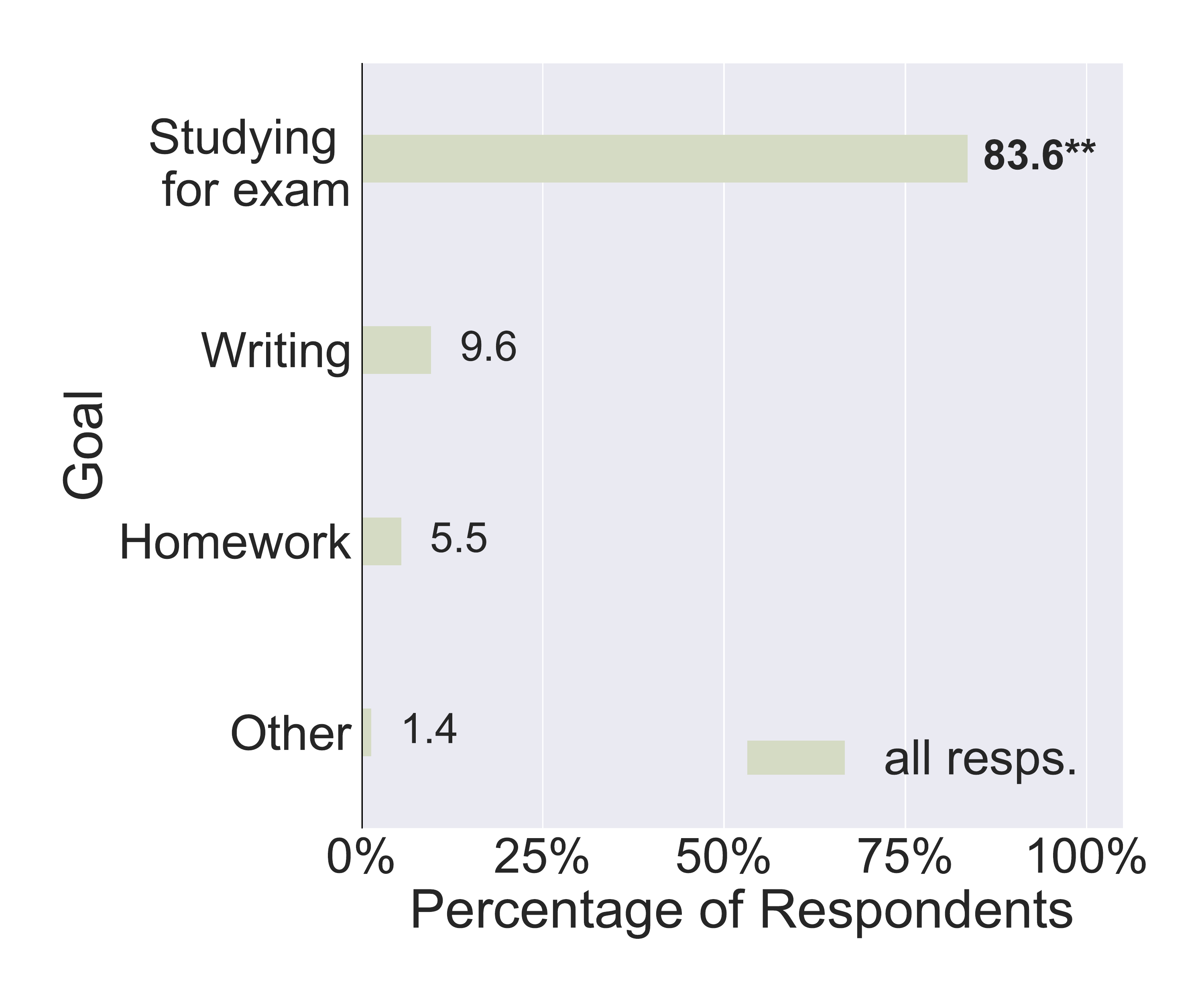}
      \label{fig:purpose_situation_goal}
  \end{subfigure}\quad
  \begin{subfigure}[b]{0.31\textwidth}
      \centering
      \caption{\textit{Situation (2):} Who made this pre-made summary? (MC, Only if \remembered{}) \\}
      \vspace{-2pt}
      \includegraphics[clip,trim=0mm 0mm 0mm 10mm,width=\textwidth]{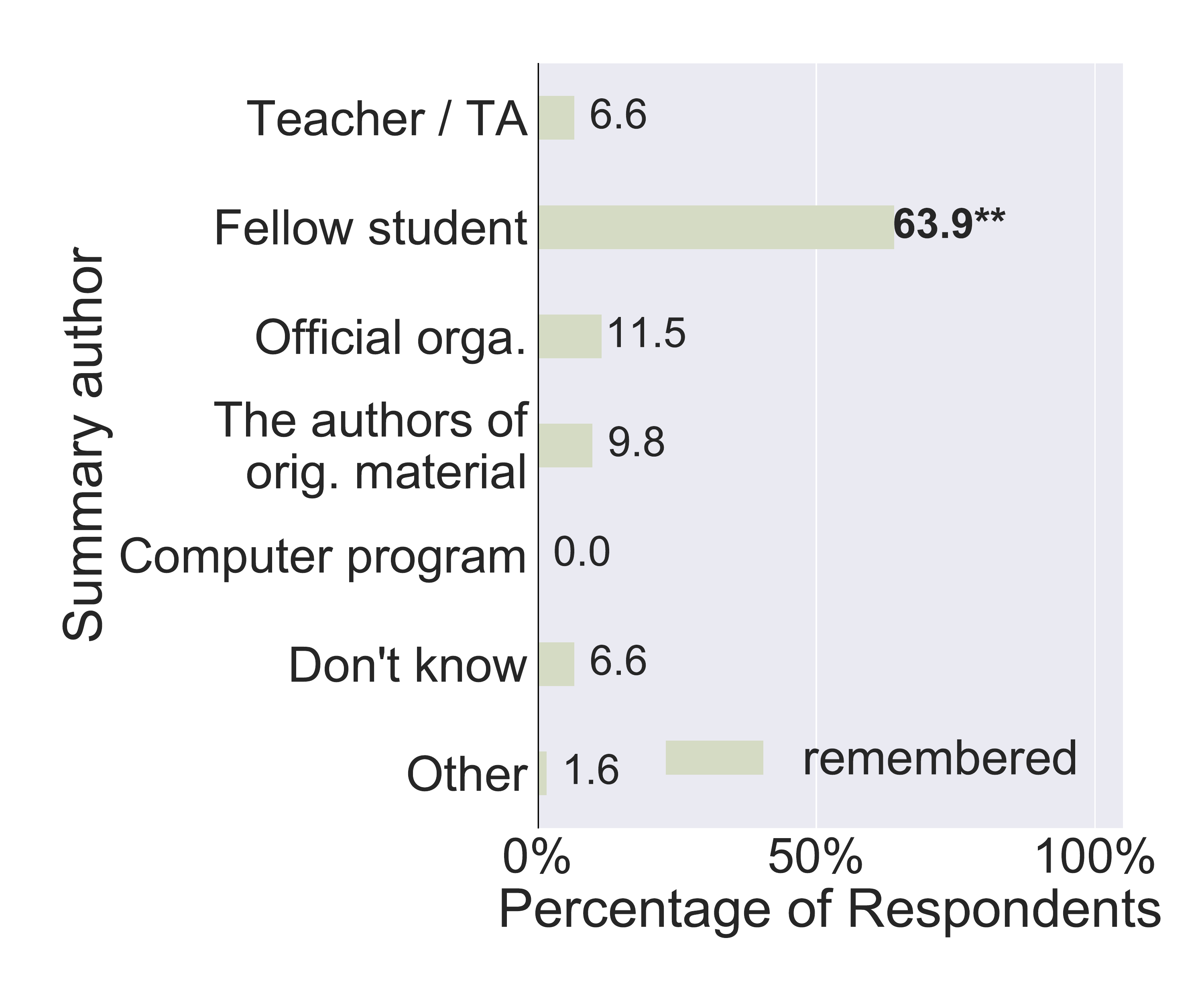}
      \label{fig:purpose_situation_who}
  \end{subfigure}\quad
  \begin{subfigure}[b]{0.31\textwidth}
      \centering
      \caption{\textit{Situation (3):} The summary was made specifically to help me (and potentially my fellow students) with my study activity (LS, Only if \remembered{})}
      \vspace{-2pt}
      \includegraphics[clip,trim=0mm 0mm 0mm 10mm,width=\textwidth]{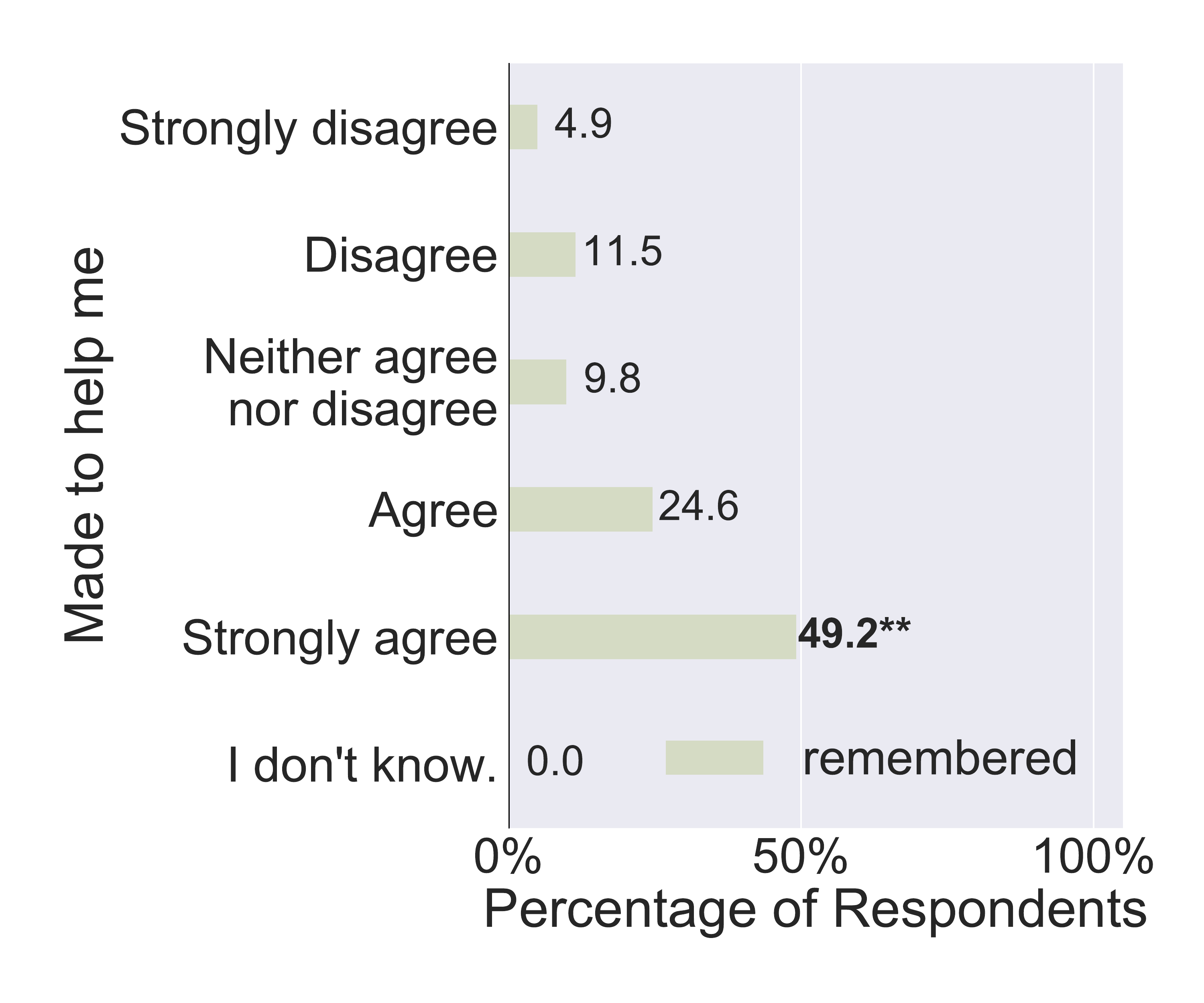}
      \label{fig:purpose_situation_help}
   \end{subfigure}
  \vspace{-1.5em}

  \begin{subfigure}[b]{0.31\textwidth}
      \centering
      \caption{\textit{Audience:} For what type of people was the summary intended? (LS) \\}
      \vspace{-2pt}
      \includegraphics[clip,trim=0mm 0mm 0mm 10mm,width=\textwidth]{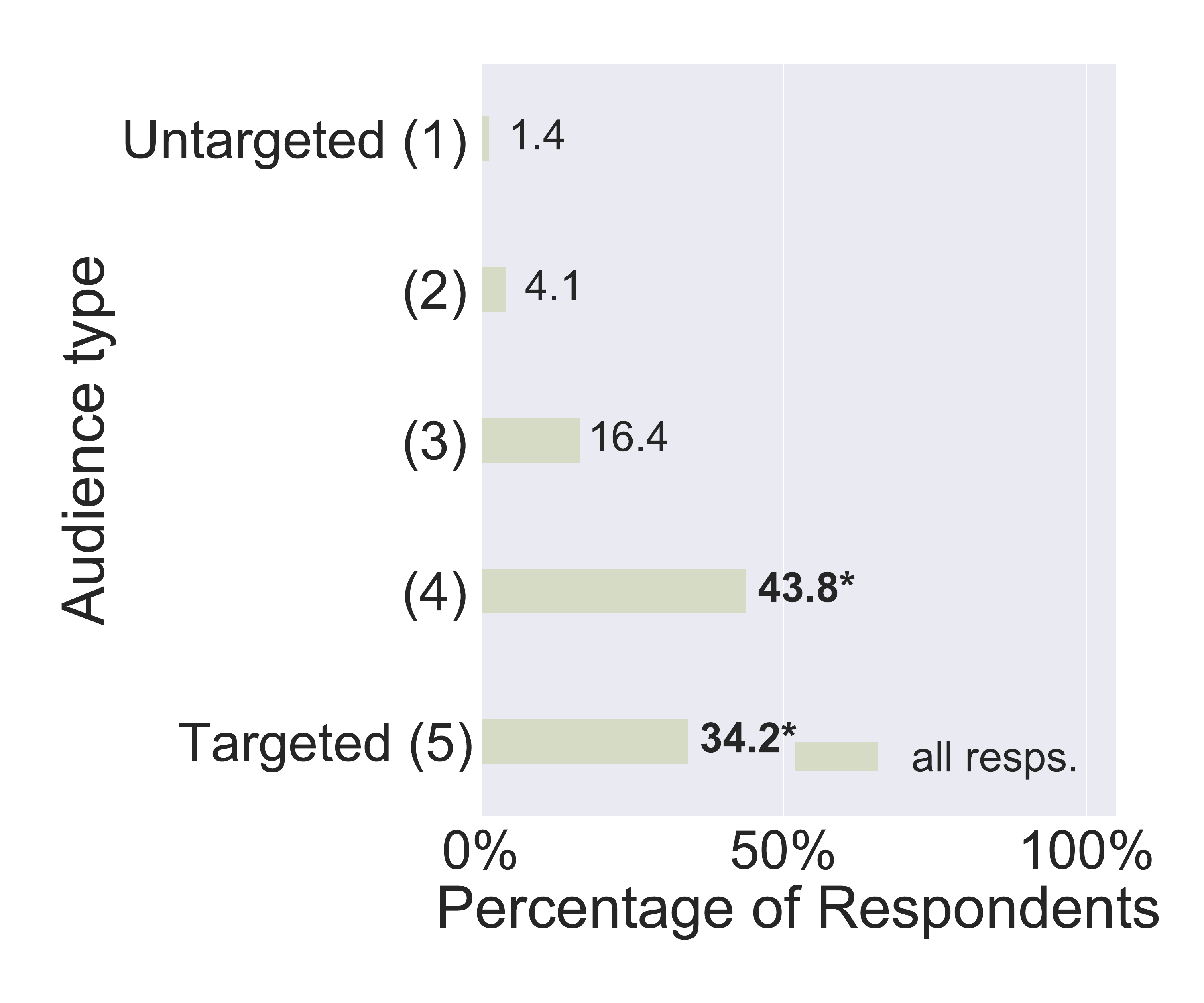}
      \label{fig:purpose_situation_audience}
  \end{subfigure}\quad
  \begin{subfigure}[b]{0.31\textwidth}
      \centering
      \caption{\textit{Use (1):} How did this summary help you with your task? (MR) \\}
      \vspace{-2pt}
      \includegraphics[clip,trim=0mm 0mm 0mm 10mm,width=\textwidth]{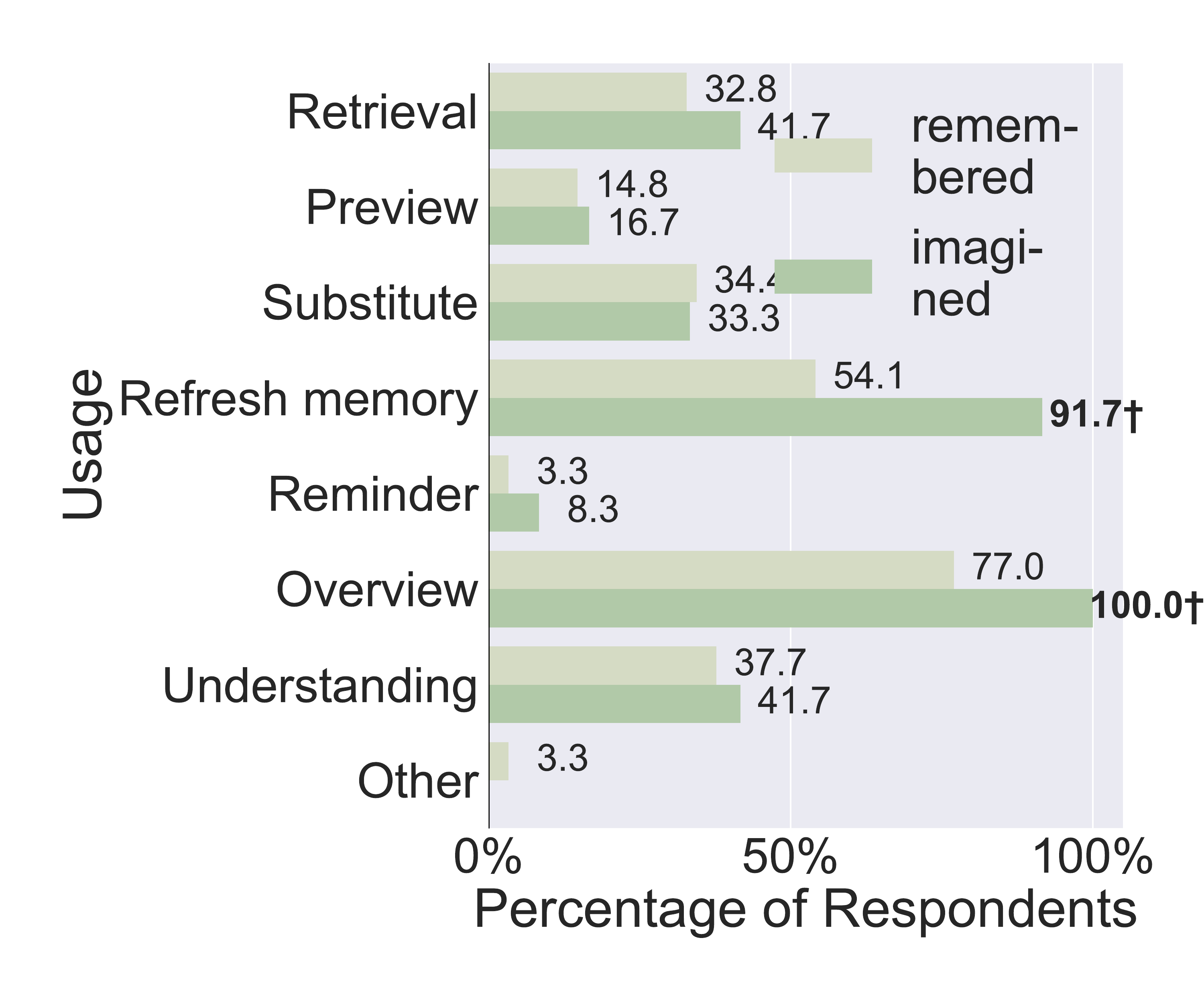}
      \label{fig:purpose_use_help_desc_im}
  \end{subfigure}\quad
  \begin{subfigure}[b]{0.31\textwidth}
      \centering
      \caption{\textit{Use (2):} Overall, how helpful was the pre-made summary for you? (LS, Only if \remembered{})}
      \vspace{-2pt}
      \includegraphics[clip,trim=0mm 0mm 0mm 10mm,width=\textwidth]{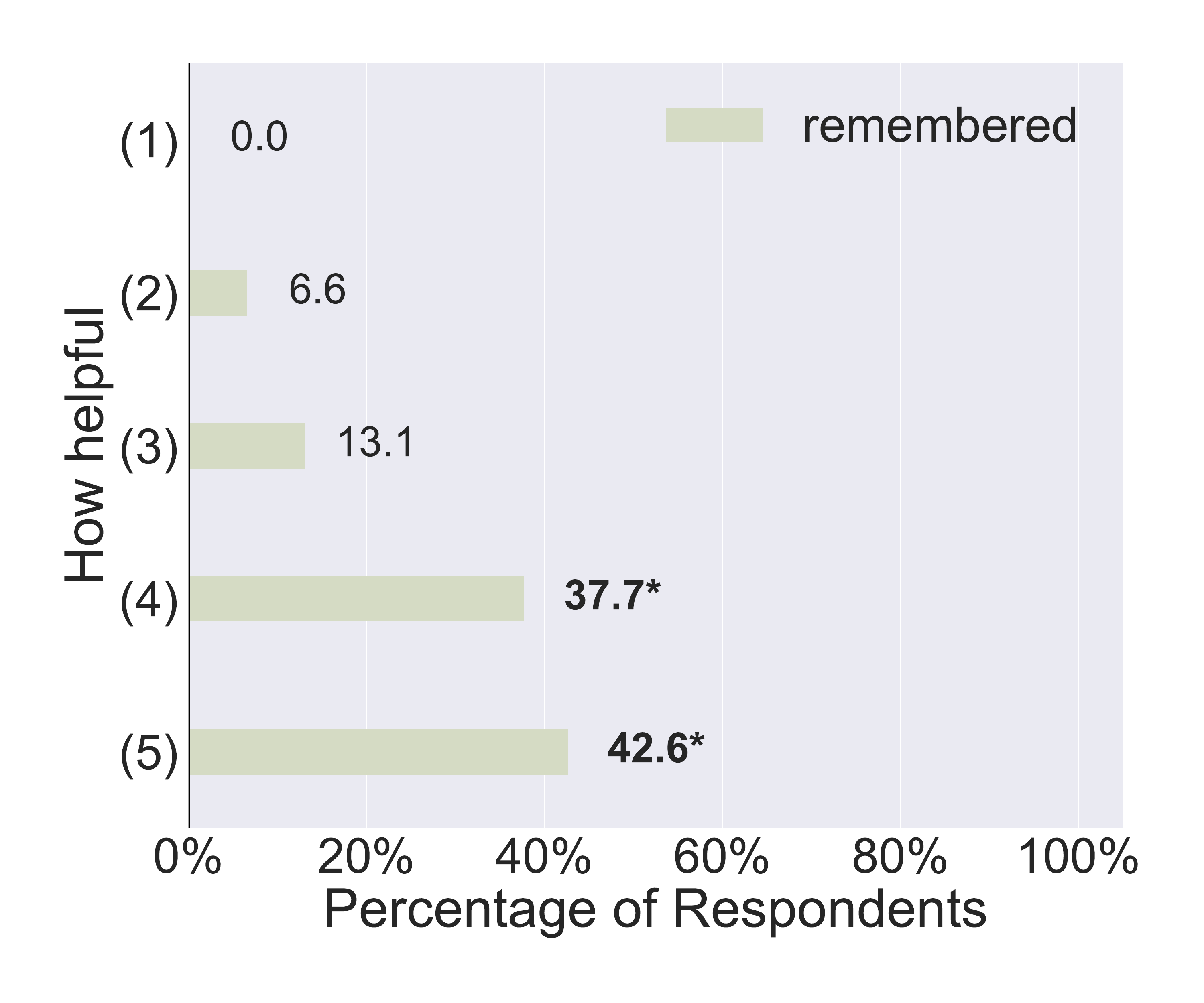}
      \label{fig:purpose_use_helpful}
   \end{subfigure}\quad
\vspace{-3em}
\caption{Results for the \textbf{\emph{purpose factor}} questions. Specific purpose factor in italics. Answer type in brackets: MC = Multiple Choice, MR = Multiple Response, LS = Likert Scale. \textbf{**} indicates significance ($\chi^2$), after Bonferroni correction, with $p\ll 0.001$, \textbf{*} with $p<0.05$. $\dagger$ indicates noteworthy results where significance was lost after correction for the number of tests. If two options are flagged, these options are not significantly different from each other, yet both were chosen significantly more often than the other options.}
\label{fig:purpose_factors}
\end{figure*}


\section{Results}
\label{sec:results}

For each question we examine the outcomes of all respondents together and of different subgroups (Table~\ref{tab:levels_of_investigation}). For space and clarity reasons, we present the results of all respondents together, unless interesting differences between groups are found. We use the question formulations as used for the \remembered{} group and abbreviate answer options. 
Answers to multiple choice and multiple response questions are presented in an aggregated manner and we ensure that none of the open answers can be used to identify individual participants.

\subsection{Identifying branches}
Of our participants, $78.0\%$ were led to the \remembered{} branch and of the remaining $22.0\%$, $78.2\%$ were led to the \imagined{} branch. We asked the few remaining participants why they could not think of a case where a pre-made summary could be useful for them. People answered that they would not trust such a summary and that making a summary themselves helped with their study activities.

\subsection{Input factors}
\label{sec:results_input_factors}

Figure~\ref{fig:input_factors} shows the input factor results. We highlight some
here. Textual input is significantly more popular than other input types (Figure~\ref{fig:input_medium}),\footnote{This is based on people's initial responses and not on the follow up question if they selected another option than `text'.} stressing the relevance of automatic text summarization. People described a diverse input for \textit{scale} and \textit{unit} (Figure~\ref{fig:input_scale_unit}), much more than the classical focus of automatic summarization suggests. Most input had a considerable amount of structure (Figure~\ref{fig:input_structure}). Structure is often discarded in automatic summarization, although it can be very informative.

\begin{table}[tp]
    \begin{tabularx}{\columnwidth}{lX}
    \toprule
      1 & All respondents together \\
      2 & \remembered{} branch vs \imagined{} branch \\
      3 & Different study fields \\
      4 & Different study levels \\
      5 & Different levels of how helpful the summary was according to participants, rated on a $5$-point Likert scale (note that only the \emph{remembered} group answered this question) \\
        \bottomrule
    \end{tabularx}
    \caption{Levels of investigation. We did not find significant differences for each, but add all for completeness.}
    \label{tab:levels_of_investigation}
    \end{table}

\subsection{Purpose factors}
\label{sec:results_purpose_factors}

\begin{figure}[!t]
  \centering
   \begin{subfigure}[b]{0.31\textwidth}
       \centering
       \caption[t]{\textit{Format (1):} What was the type of the summary? (MC)}
       \vspace{-2pt}
       \includegraphics[clip,trim=0mm 0mm 0mm 10mm,width=\textwidth]{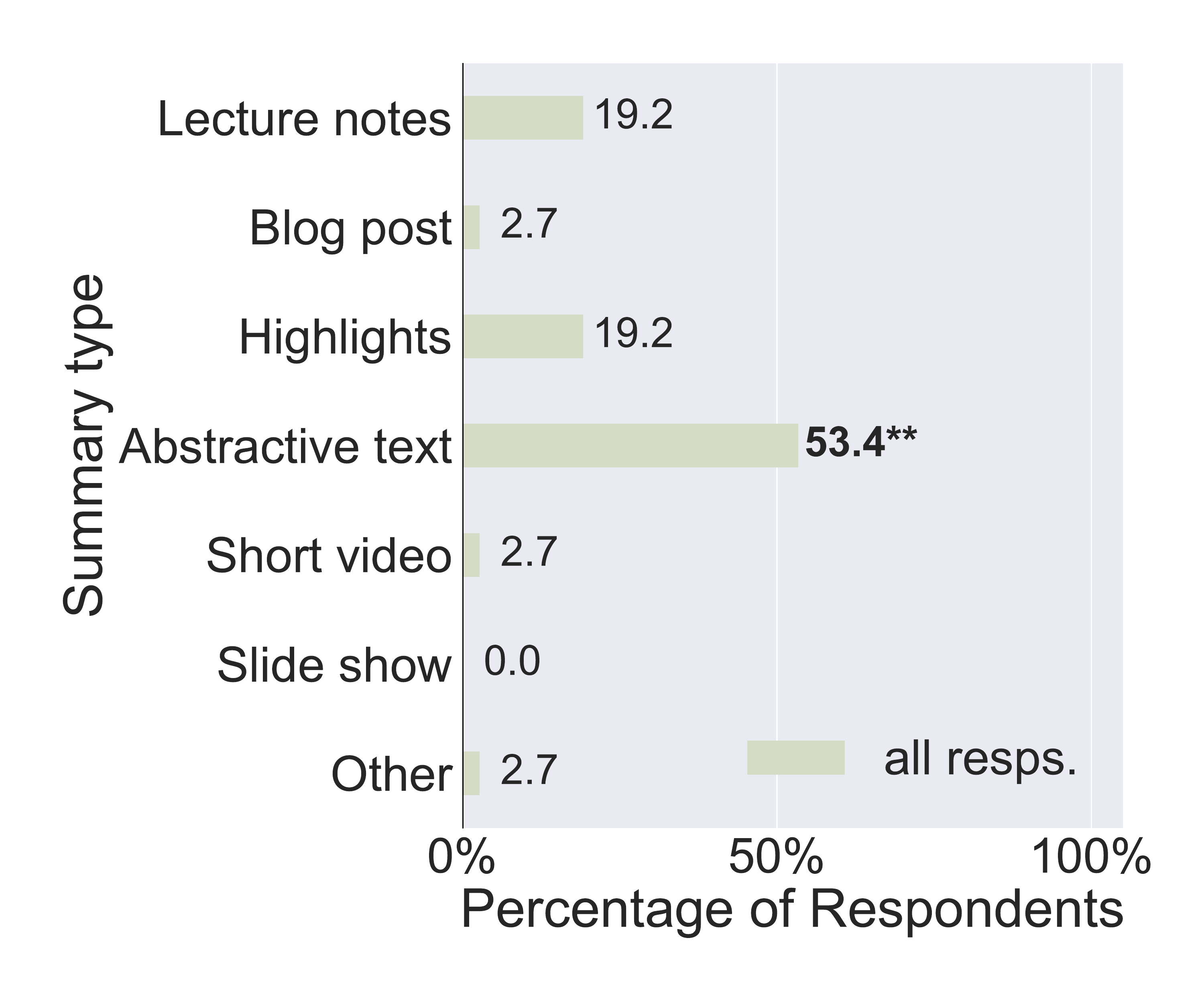}
       \label{fig:output_format_type}
       \vspace{-1.5em}
   \end{subfigure}\quad%

   \begin{subfigure}[b]{0.31\textwidth}
       \centering
       \caption[t]{\textit{Format (2):} How was the summary structured? (MR)}
       \vspace{-2pt}
       \includegraphics[clip,trim=0mm 0mm 0mm 10mm,width=\textwidth]{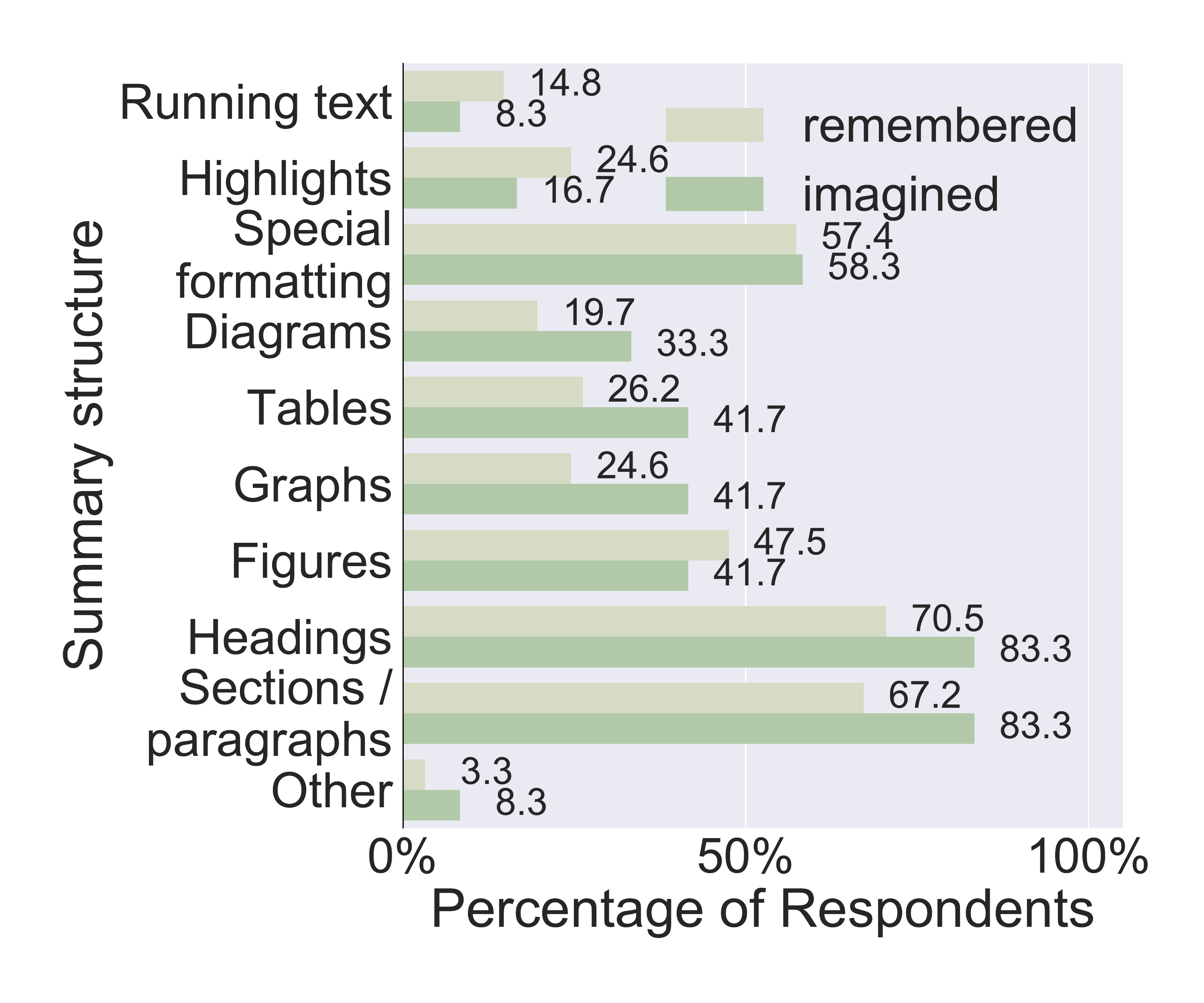}
       \label{fig:output_format_structure}
   \end{subfigure}\quad%

   \vspace{-1.5em}

   \begin{subfigure}[b]{0.31\textwidth}
       \centering
       \caption[t]{\textit{Material:} How much of the study material was covered by the summary? (LS)}
       \vspace{-2pt}
       \includegraphics[clip,trim=0mm 0mm 0mm 10mm,width=\textwidth]{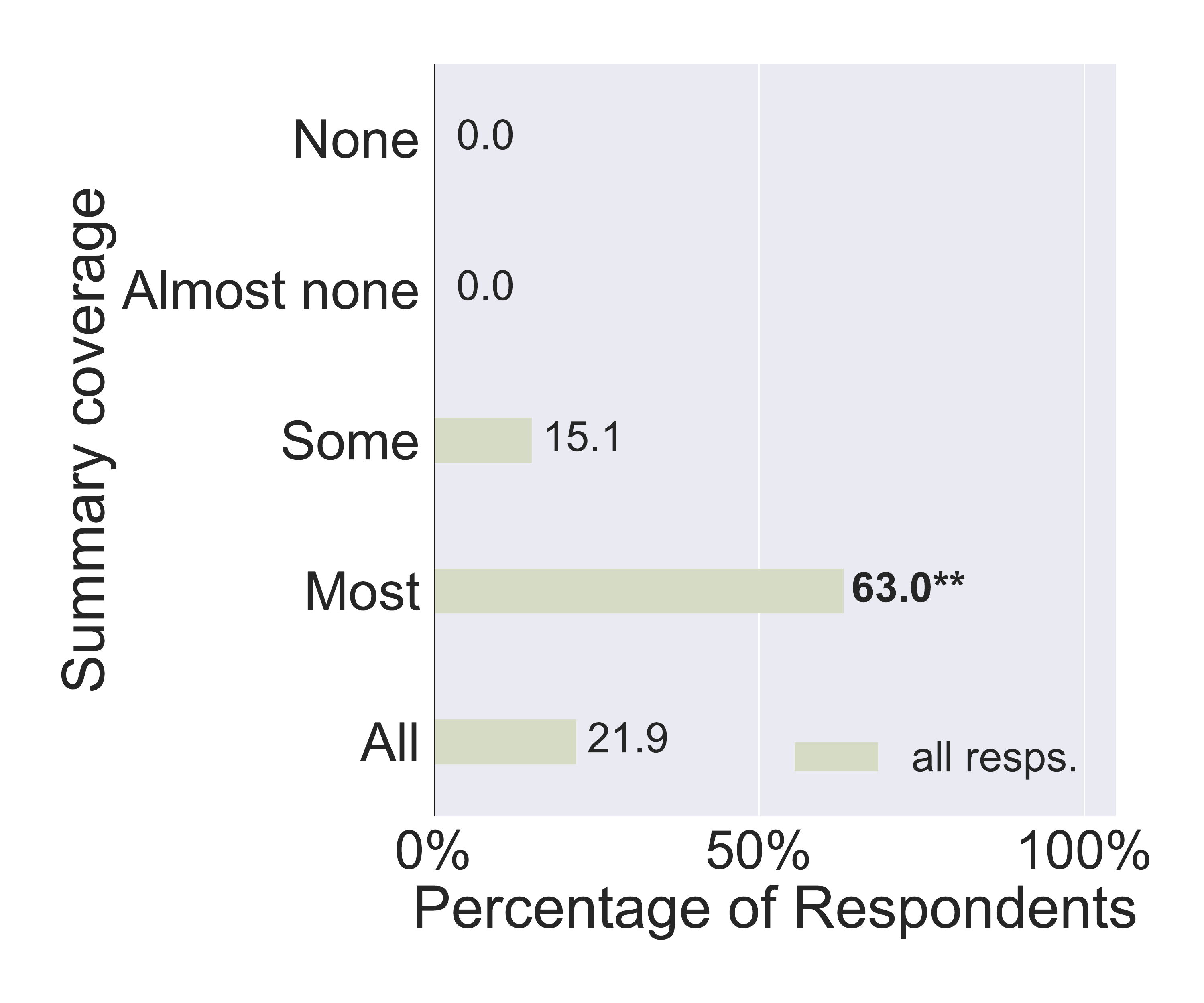}
       \label{fig:output_material_combined}
      \vspace{-1.5em}
   \end{subfigure}\quad
   \begin{subfigure}[b]{0.31\textwidth}
       \centering
       \caption[t]{\textit{Style:} What was the style of this summary? (MC)}
       \vspace{-2pt}
       \includegraphics[clip,trim=0mm 0mm 0mm 10mm,width=\textwidth]{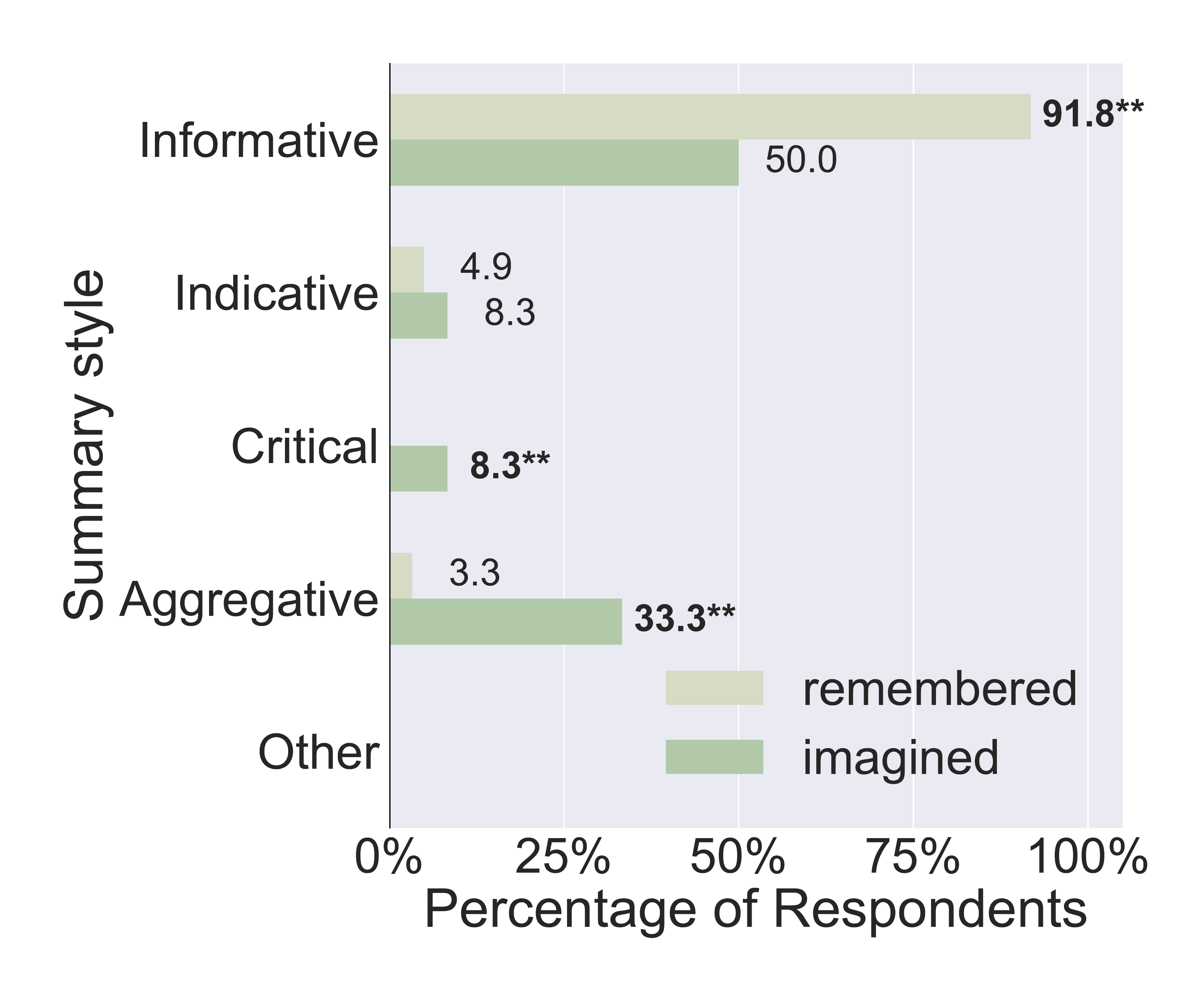}
       \label{fig:output_style}
   \end{subfigure}\quad
\vspace{-1.5em}
\caption{Results for the \textbf{\emph{output factor}} questions. Specific output factor in italics. Answer type in brackets: MC = Multiple Choice, MR = Multiple Response, LS = Likert Scale. \textbf{**} indicates significance ($\chi^2$ or Fisher's exact test), after Bonferroni correction, with $p\ll 0.001$, \textbf{*} with $p<0.05$.
}
\label{fig:output_factors}
\end{figure}

Figure~\ref{fig:purpose_factors} shows the purpose factor results. Participants indicated that the summary was \textit{helpful} or \textit{very helpful} (Figure~\ref{fig:purpose_use_helpful}), which allows us to draw valid conclusions from the survey.\footnote{Because we do not find significant differences in the overall results when we exclude the few participants who did not find their summary helpful and we do not find many correlations w.r.t.\ how helpful a summary was and a particular context factor, we include all participants in the analysis, regardless of how helpful they found their summary, for completeness.} 
We now highlight some results from the other questions in this category. For the intended \textit{audience} of the summaries, students selected level (4) and (5) (\textit{``a lot (4) or full (5) domain knowledge is expected from the users of the summary"}) significantly more often than the other options (Figure~\ref{fig:purpose_situation_audience}). Although perhaps unsurprising given our target group, it is an important outcome as this requires a different level of detail than, for example, a brief overview of a news article. 
People used the summaries for many different use-cases (Figure~\ref{fig:purpose_use_help_desc_im}), whereas current research on automatic summarization mainly focuses on giving an overview of the input. We show the results for the \remembered{} vs. \imagined{} splits, as the \imagined{} group chose \textit{refresh memory} and \textit{overview} more often than the \remembered{} group (Fisher's exact test, $p<0.05$). Although not significant after a Bonferroni correction, this can still be insightful for future research directions.
Lastly, participants in the \imagined{} group ticked more boxes than participants in the \remembered{} group: $3.33$ vs.\ $2.57$ per participant on average, stressing the importance of considering many different use-cases for automatically generated summaries.

\subsection{Output factors}

Figure~\ref{fig:output_factors} shows the results for the output factor questions. Textual summaries were significantly more popular than other summary types (Figure~\ref{fig:output_format_type}), which again stresses the importance of automatic text summarization. Most participants indicated that the summary covered (or should cover) most of the input \textit{material} (Figure~\ref{fig:output_material_combined}). For the output factor \textit{style} we find an interesting difference between the \remembered{} and \imagined{} group (Figure~\ref{fig:output_style}). Whereas the \remembered{} group described significantly more often an \textit{informative} summary, the \imagined{} group opted significantly more often for a \textit{critical} or \textit{aggregative} summary. Most research on automatic summarization focusses on informative summaries only. For the output factor \textit{structure} (Figure~\ref{fig:output_format_structure}), people described a substantially richer format of the pre-made summaries than adopted in most research on automatic summarization. Instead of simply a running text, the vast majority of people indicated that the summary contained (or should contain) structural elements such as special formatting, diagrams, headings, etc. Moreover, 
the \imagined{} group ticked more answer boxes on average than the \remembered{} group: $4.17$ vs.\ $3.56$ per participant, indicating a desire for structure in the generated summaries, which is supported by the open answer questions.

\noindent \textbf{Open answer questions.}
We asked participants in the \remembered{} group how the summary could be transformed into their ideal summary and $86.9\%$ of these participants made suggestions. Many of those include adding additional structural elements to the summary, like figures, tables or structure in the summary text itself. For example, one of the participants wrote: \textit{``An ideal summary is good enough to fully replace the original (often longer) texts contained in articles that need to be read for exams. The main purpose behind this is speed of learning from my experience. More tables, graphs and visual representations of the study material and key concepts / links would improve the summary, as I would faster comprehend the study material.''}
Another participant wrote: \textit{``-- colors and a key for color-coding -- different sections, such as definitions on the left maybe and then the rest of the page reflects the structure of the course material with notes on the readings that have many headings and subheadings.''}

Another theme is the desire to have more examples in the summary. One participant wrote: \textit{``More examples i think. For me personally i need examples to understand the material. Now i needed to imagine them myself''}.  

Some participants wrote that they would like a more personalized summary, for example:  \textit{``I'd highlight some things I find difficult. So I'd personalise the summary more.''} Another participant wrote: \textit{``Make it more personalized may be. These notes were by another student. I might have focussed more on some parts and less on others.''} 

\subsection{Trustworthiness and future features}

Of all participants, $48.0\%$ indicated that it would not make a difference to them whether a summary is machine- or human-generated, as long as the quality is as good as a human-generated one. This last point is reflected in which types of summaries participants would trust more. People opted significantly more often for a human-generated one. For the future feature questions, \textit{adding more details to the summary} and \textit{answering questions based on the content of the summary} were very popular. We give a full account in Appendix~\ref{sec:appendix_future_features}.


\section{Implications and Perspectives}
\label{sec:implications}

\subsection{Future research directions}

Our findings have important implications for the design and development of future automatic summarization methods. We present these in Table~\ref{tab:implications}, per context factor. Summarizing, the research developments as summarized in Section~\ref{sec:related_work} are encouraging, yet given that automatic summarization methods increasingly mediate people's lives, we argue that more attention should be devoted to its stakeholders, i.e., to the purpose factors. Here we have shown that students, an important stakeholder group, have different expectations of pre-made summaries than what most automatic summarization methods offer. These differences include the type of input material that is to be summarized, but also how these summaries are presented. Presumably, this also holds for other stakeholder groups and thus we hope to see our survey used for different target groups in the future.

\noindent \textbf{Datasets.} To support these future directions we need to expand efforts on using and collecting a wide variety of datasets. Most recent data collection efforts are facilitating different input factors -- the purpose and output factors need more emphasis.

\begin{table}[t]
\begin{tabularx}{\columnwidth}{X}
\toprule
  \textbf{Input Factors} \\
  \midrule
  \begin{minipage}[t]{\linewidth}
  Stronger focus on developing methods that can:
  \begin{itemize}[leftmargin=*,nosep]
  \item handle a wide variety and a mixture of \textbf{different types} of input documents at once;
  \item understand the \textbf{relationships} between different input documents;
  \item use the \textbf{structure} of the input document(s).
  \end{itemize}
  \end{minipage}
  \\
  \midrule
	\textbf{Purpose Factors} \\
  \midrule
  \begin{minipage}[t]{\linewidth}
  \begin{itemize}[leftmargin=*,nosep]
  \item Explicitly define a \textbf{standpoint} on the purpose factors in each research project;
  \item Include a comprehensive \textbf{evaluation} methodology to evaluate usefulness. We propose this in Section~\ref{sec:usefulness_evaluation}.
  \end{itemize}
  \end{minipage}
  \\
  \midrule
	\textbf{Output Factors} \\
  \midrule
  Stronger focus on developing methods that can:
  \begin{minipage}[t]{\linewidth}
  \begin{itemize}[leftmargin=*,nosep]
  \item output different summary \textbf{styles}, e.g., informative, aggregative or critical. Especially the last two require a \textbf{deeper understanding} of the input material than current models have;
  \item explicitly model and understand \textbf{relationships} between different elements in the summary and potentially relate this back to the input document(s).
  \end{itemize}
  \end{minipage} \\
	\bottomrule
\end{tabularx}
\caption{Implications for future research directions.}
\label{tab:implications}
\end{table}

\noindent Our findings also impact the evaluation of summarization methods. We discuss this next.

\subsection{Usefulness as evaluation methodology}
\label{sec:usefulness_evaluation}

Following~\citet{jones1999automatic} and \citet{mani2001automatic}, we argue that
a good choice of context factors is crucial in producing useful summaries for users. 
It is important to explicitly evaluate this. The few existing methods to evaluate usefulness are very resource demanding~\cite[e.g.,][]{riccardi2015sensei} or not comprehensive enough~\cite[e.g.,][]{duc2003, dorr-etal-2005-methodology}. Thus, we propose a feasible and comprehensive method to evaluate usefulness.

For the evaluation methodology, we again use the context factors. Before the design and development of the summarization method the intended purpose factors need to be defined. Especially the fine-grained factor \textit{use} is important here. Next, the output factors need to be evaluated on the use factors. For this, we take inspiration from research on simulated work tasks~\cite{borlund2003iir}. Evaluators should be given a specific task to imagine, e.g., \textit{writing a news article}, or \textit{studying for an exam}. This task should be relatable to the evaluators, so that reliable answers can be obtained~\cite{borlund2016study}. With this task in mind, evaluators should be asked to judge two summaries in a pairwise manner on their usefulness, in the following format:  \textit{The [output factor] of which of these two summaries is most useful to you to [use factor]?} For example: \textit{The style of which of these two summaries is most useful to you to substitute a chapter that you need to learn for your exam preparation?}
It is critical to ensure that judges understand the meaning of each of the evaluation criteria -- \textit{style} and \textit{substitute} in the example.
We provide example questions for each of the use and output factors in Appendix~\ref{sec:appendix_evaluation_questions}.

\section{Conclusion}
\label{sec:conclusion}
In this paper we focused on users of automatically generated summaries and argued for a stronger emphasis on their needs in the design, development and evaluation of automatic summarization methods. We led by example and proposed a survey methodology to identify these needs. Our survey is deeply grounded in past work by~\citet{jones1999automatic} on context factors for automatic summarization and can be re-used to investigate a wide variety of users. In this work we use our survey to investigate the needs of university students, an important target group of automatically generated summaries. We found that the needs identified by our participants are not fully supported by current automatic summarization methods and we proposed future research directions to accommodate these needs. Finally, we proposed an evaluation methodology to evaluate the usefulness of automatically generated summaries.

\section{Ethical Impact}
\label{sec:ethical_impact}

With this work we hope to take a step in the right direction to make research into automatic summarization more inclusive, by explicitly taking the needs of users of these summaries into account. As stressed throughout the paper, these needs are different per user group and therefore it is critical that a wide variety of user groups will be investigated. There might also be within group differences. For example, in this work we have focussed on students from universities in one country, but students attending universities in other geographical locations and with different cultures might express different needs. It is important to take these considerations into account, to limit the risk of overfitting on a particular user group and potentially harming other user groups.

\section*{Acknowledgements}
We thank Jacobijn Sandberg and Ana Lucic for helpful comments and feedback.
This research was supported by the Nationale Politie.
All content represents the opinion of the authors, which is not necessarily shared or endorsed by their respective employers and/or sponsors.

\bibliography{anthology,custom}
\bibliographystyle{acl_natbib}

\appendix

\onecolumn
\section{Overview context factors}
\label{sec:appendix_overview_summary_factors}

\definecolor{Gray0}{gray}{0.6}
\definecolor{Gray1}{gray}{0.8}
\definecolor{Gray2}{gray}{0.95}

\renewcommand{\arraystretch}{1.3}
\begin{table}[h]
\centering%
\begin{tabularx}{\textwidth}{X|X|X}
\rowcolor{Gray1}

\textbf{Input Factors}         & \textbf{Purpose Factors}    & \textbf{Output Factors}    \\

\rowcolor{Gray2}
\textit{\textbf{Form}}         & \textit{\textbf{Situation}} & \textit{\textbf{Material}} \\

\textit{Structure:} How is the input text structured? E.g., subheadings, rhetorical patterns, etc.           & \textit{Tied:} It is known who will use the summary, for what purpose and when.               & \textit{Covering: } The summary covers all of the important information in the source text.          \\

\textit{Scale:} How large is the input data that we are summarizing? E.g., a book, a chapter, a single article, etc.                 & \textit{Floating:} It is not (exactly) known who will use the summary, for what purpose or when.           & \textit{Partial:} The summary (intentionally) covers only parts of the important information in the source text.            \\

\textit{Medium:} What is the input language type? E.g., full text, telegraphese style, etc. This also refers to which natural language is used.               & \textit{\textbf{Audience}} \cellcolor{Gray2}  & \textit{\textbf{Format}} \cellcolor{Gray2}   \\

\textit{Genre:} What type of literacy does the input text have? E.g., description, narrative, etc.                 & \textit{Targetted:} A lot of domain knowledge is expected from the readers of the summary.           & \textit{Running:} The summary is formatted as an abstract like text. \\

\textit{\textbf{Subject Type}} \cellcolor{Gray2} & \textit{Untargetted:} No domain knowledge is expected from the readers of the summary.        & \textit{Headed:} The summary is structured following a certain standardised format with headings and other explicit structure.  \\

\textit{Ordinary:} Everyone could understand this input type.             & \textit{\textbf{Use}}  \cellcolor{Gray2}      & \textit{\textbf{Style}} \cellcolor{Gray2}   \\

\textit{Specialized:} You need to speak the jargon to understand this input type.           & \textit{Retrieving:} Use the summary to retrieve source text.         & \textit{Informative: } The summary conveys the raw information that is in the source text.     \\

\textit{Restricted}: The input type text is only understandable for people familiar with a certain area,  for example because it contains local names.            & \textit{Previewing:} Use the summary to preview a text.         & \textit{Indicative:} The summary just states the topic of the source text, nothing more.         \\

\textit{\textbf{Unit}} \cellcolor{Gray2}        & \textit{Substitutes:} Use the summary to substitute the source text.        & \textit{Critical:} The summary gives a critical review of the merits of the source text.          \\

\textit{Single:} Only one input source is given.               & \textit{Refreshing:} Use the summary to refresh ones memory of the source text.         & \textit{Aggregative:} Different source texts are put in relation to one another to give an overview of a certain topic.       \\

\textit{Multi:} Multiple input sources are given.                 & \textit{Prompts:} Use the summary as action prompt to read the source text.            &
\end{tabularx}%
\caption{Overview of different context factors classes defined by~\citet{jones1999automatic}, with descriptions of the factors within these classes.}
\label{tab:overview_context_factors}
\vspace*{-5cm}
\end{table}


\onecolumn

\section{Survey overview}
\label{sec:appendix_survey_overview}

\begin{figure}[h]
  \centering
\includegraphics[width=0.66\textwidth]{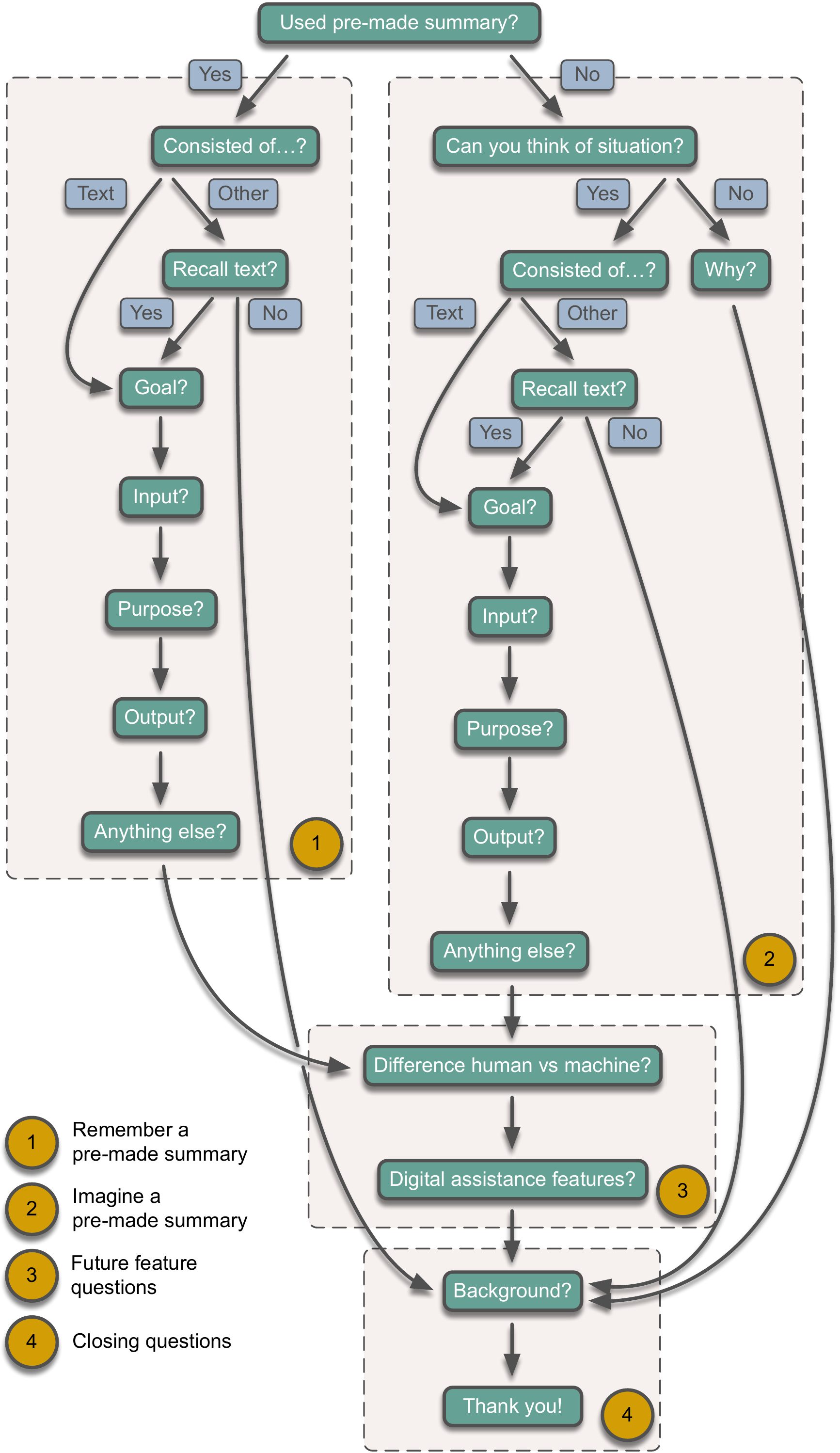}
\caption{Overview survey design.}
\label{fig:overview_survey}
\end{figure}


\section{Verbatim survey overview}
\label{sec:appendix_literal_survey_overview}

\addtocounter{footnote}{1}
\footnotetext{\url{http://surveymonkey.com}}
\addtocounter{footnote}{-1}

\tablehead{
\toprule
\textbf{Question Nr.} & \textbf{Question and Answer Options} \\
\midrule
}
\captionof{table}{A complete overview of the survey. This table includes the explanation that participants received, as well as all the questions and the answer options. If a question was the start of a branch, the direction of the branch has been written behind the answer options in italic. (This was never shown to the participants.) Note that the survey was performed in SurveyMonkey.\protect\footnote{\protect\url{http://surveymonkey.com}} The survey had a lay-out as provided by SurveyMonkey, i.e., it consisted of different pages and colors were used to highlight certain important parts in texts.}
\begin{xtabular}{@{\,}l <{\hskip 2pt} >{\raggedright\arraybackslash}p{0.82\textwidth}}

    Q1  & \textbf{Introduction and Instructions} \\

        & Thank you for taking the time to fill out this survey! Before you start, please take the time to read these instructions carefully. If you still have any questions after reading the instructions, please send them to {\small \url{m.a.terhoeve@uva.nl}}. \\
        & We will give away 10 {\small \url{bol.com}} vouchers of 10 euros each among the participants. If you would like to take part in the raffle, you can leave your email address at the end of this survey. \\

        & \textbf{Goal of the study} \\

        & The goal of this survey is to get insight in how summaries help or can help you when studying. \\

        & \textbf{What the survey will look like} \\

        & In what follows you will get questions that aim to develop an understanding for:
        \begin{itemize}[leftmargin=*]
          \item For which types of study material it is useful to have summaries
          \item How these summaries can help you with your task
          \item What these summaries should look like
        \end{itemize} \\

        & We expect this survey to take approximately 10 minutes of your time. \\

        &  Use the next button to go to the next page once you have filled out all the questions on the page. Use the prev button to go back one page. \\

        & \textbf{About your privacy} \\

        & We value your privacy and will process your answers anonymously. The answers of all participants in this survey will be used to gain insight in how pre-made summaries can be helpful for different types of studying activities. The answers will be presented in a research paper about this topic. This will be done either in an aggregated manner, or by citing verbatim examples of the answers. Again, this will all be done anonymously. \\

        & \\

        & \textbf{I agree that I have read and understood the instructions. I also understand that my participation in this survey is voluntarily.}
        \begin{itemize}[label=$\square$, leftmargin=*]
            \item I agree
        \end{itemize} \\

    Q2  & \textbf{Important! Some background knowledge you need to know} \\

        & Throughout this survey we make use of the term pre-made summary. It is very important that you understand what this means. On this page we explain this term, so please make sure to read this carefully. \\

        & \textbf{Definition pre-made summary} \\

        & One type of summary is one that you make yourself. Another type of summary is one that has been made for you. In this survey, we focus on this latter type and we call them pre-made summaries. \\

        & \textbf{Who makes these pre-made summaries?} \\

        & These pre-made summaries can be made by a person, for example your teacher, your friend, a fellow student or someone at some official organisation, etc. The pre-made summaries can also be made by a computer. \\

        & \textbf{What kinds of summaries are we talking about?} \\

        & There are no restrictions on what these pre-made summaries can look like. On the contrary, that is one of the things we aim to find out with this survey! But, to give some examples, you could think of a written overview of a text book, highlights in text to draw your attention to important bits, blog posts, etc. These are really just examples and don't let them limit your creativity! You can come up with any example of a pre-made summary that is helpful for you. \\

        & \\

        & \textbf{Yes, I understand what a pre-made summary is!}
          \begin{itemize}[label=$\square$, leftmargin=*, nosep]
            \item Yes
          \end{itemize} \\

        Q3 & Please think back to your recent study activities. Examples of study activities can be: studying for an exam, writing a paper, doing homework exercises, etc. Note that these are just examples, any other study activity is fine too. \\

        & \textbf{Did you use a pre-made summary in any of these study activities?}
        \begin{itemize}[label=$\square$, leftmargin=*, nosep]
          \item Yes \textit{-- participants are led to Q6}
          \item No \textit{-- participants are led to Q4}
        \end{itemize} \\

        Q4 & \textbf{Can you think of one of your recent study activities where a pre-made summary would have been useful for you?}
        \begin{itemize}[label=$\square$, leftmargin=*, nosep]
          \item Yes \textit{-- participants are led to Q25}
          \item No \textit{-- participants are led to Q5}
        \end{itemize}

        \\

        Q5 & \textbf{Why do you think a pre-made summary would not have helped you with any of your recent study activities? }\\

        & Open response \textit{-- participants are led to Q48}

        \\ \midrule

        \multicolumn{2}{l}{\textbf{Start branch of participants who described an existing summary}} \\ \midrule

        & If you have multiple study activities where you used a pre-made summary, please take the one where you found the pre-made summary most useful. \\

        Q6 & \textbf{The original study material consisted of}
        \begin{itemize}[label=$\square$, leftmargin=*, nosep]
          \item Mainly text \textit{-- participants are led to Q8}
          \item Mainly figures \textit{-- participants are led to Q7}
          \item Mainly video \textit{-- participants are led to Q7}
          \item Mainly audio \textit{-- participants are led to Q7}
          \item A combination of some or all of the above \textit{-- participants are led to Q7}
          \item I do not know, because I have not seen the study material \textit{-- participants are led to Q7}
          \item Other (please specify) \textit{-- participants are led to Q7}
        \end{itemize}

        \\

        Q7 & \textbf{For now we narrow down our survey to study material that is mostly textual. Do you recall any other recent study activity where you made use of a pre-made summary and where the original study material mainly consisted of text?}
        \begin{itemize}[label=$\square$, leftmargin=*, nosep]
          \item Yes \textit{-- participants are led to Q8}
          \item No \textit{-- participants are led to Q48}
        \end{itemize}

        \\

        Q8 & \textbf{What was the goal of this study activity?}
        \begin{itemize}[label=$\square$, leftmargin=*, nosep]
          \item Studying for an exam
          \item Writing a paper / essay / report / etc.
          \item Doing homework exercises
          \item Other (please specify)
        \end{itemize}

        \\

        Q9 & \textbf{Who made this pre-made summary?}
        \begin{itemize}[label=$\square$, leftmargin=*, nosep]
          \item A teacher or teaching assistant
          \item A fellow student
          \item An official organisation
          \item The authors of the original study material
          \item A computer program
          \item I am not sure, I found it online
          \item Other (please specify)
        \end{itemize}

        \\

        & Now some questions will follow about what the study material that was summarized looked like. \\

        Q10 & \textbf{What was the length of the study material?}
        \begin{itemize}[label=$\square$, leftmargin=*, nosep]
          \item A single article
          \item Multiple articles
          \item A single book chapter
          \item Multiple book chapters from the same book
          \item Multiple book chapters from various books
          \item A combination of the above
          \item I do not know because I have not seen the study material, only the summary
          \item Other (please specify)
        \end{itemize}

        \\

        Q11 & \textbf{How was the study material structured? (Multiple answers possible)}
        \begin{itemize}[label=$\square$, leftmargin=*, nosep]
          \item There was no particular structure - e.g. just one large text
          \item The text contained a title or titles
          \item The text contained subheadings
          \item The text consisted of different chapters
          \item The text consisted of different sections and / or paragraphs
          \item I do not know because I have not seen the study material, only the summary
          \item Other (please specify)
        \end{itemize}

        \\

        Q12 & \textbf{What was the genre of the study material?}
        \begin{itemize}[label=$\square$, leftmargin=*, nosep]
          \item Mainly educational (such as a text book (chapter))
          \item Mainly scientific (such as an academic article, publication, report, etc)
          \item Mainly nonfiction writing (such as (auto)biographies, history books, etc)
          \item Mainly fiction writing (such as novels, short fictional stories, etc)
          \item Other (please specify)
        \end{itemize}

        \\

        Q13 & \textbf{How would you classify the difficulty level of the study material?}
        \begin{itemize}[label=$\square$, leftmargin=*, nosep]
          \item Ordinary: most people would be able to understand it
          \item Specialized: you need to know the jargon of the field to be able to understand it
          \item Geographically based: you can only understand it if you are familiar with a certain area, for example because it contains local names
        \end{itemize}

        \\

        & Now we will ask some questions about the purpose of the pre-made summary that you used. \\

        Q14 & \textbf{The summary was made specifically to help me (and potentially fellow students) with my study activity.}

        \begin{tabularx}{0.7\columnwidth}{
          !{\hskip 2pt}>{\centering\arraybackslash}p{0.11\columnwidth}
          !{\hskip 2pt}>{\centering\arraybackslash}p{0.11\columnwidth}
          !{\hskip 2pt}>{\centering\arraybackslash}p{0.11\columnwidth}
          !{\hskip 2pt}>{\centering\arraybackslash}p{0.11\columnwidth}
          !{\hskip 2pt}>{\centering\arraybackslash}p{0.11\columnwidth}
          !{\hskip 2pt}>{\centering\arraybackslash}p{0.11\columnwidth}}

          Strongly disagree & Disagree & Neither agree nor disagree & Agree & Strongly agree & I don't know \\

          $\square$ & $\square$ & $\square$ & $\square$ & $\square$ & $\square$
        \end{tabularx}

        \\

        Q15 & \textbf{For what type of people was the summary intended? Your score can range from (1) Untargetted, to (5) Targetted.}

        \begin{tabularx}{0.7\columnwidth}{
          !{\hskip 2pt}>{\centering\arraybackslash}p{0.13\columnwidth}
          !{\hskip 2pt}>{\centering\arraybackslash}p{0.13\columnwidth}
          !{\hskip 2pt}>{\centering\arraybackslash}p{0.13\columnwidth}
          !{\hskip 2pt}>{\centering\arraybackslash}p{0.13\columnwidth}
          !{\hskip 2pt}>{\centering\arraybackslash}p{0.13\columnwidth}}

          Untargetted: No domain knowledge is expected from the users of the summmary. & & & & Targetted: Full domain knowledge is expected from the users of the summmary. \\
          (1) & (2) & (3) & (4) & (5) \\

          $\square$ & $\square$ & $\square$ & $\square$ & $\square$
        \end{tabularx}

        \\

        Q16 & \textbf{How did this summary help you with your task? (Multiple answers possible)}
        \begin{itemize}[label=$\square$, leftmargin=*, nosep]
          \item The summary helped to retrieve parts of the original study material
          \item I used the summary to preview the text that I was about to read
          \item I used the summary as a substitute for the original study material
          \item I used the summary to refresh my memory of the original study material
          \item I used the summary as a reminder that I had to read the original study material
          \item The summary helped to get an overview of the original study material
          \item The summary helped to understand the original study material
          \item Other (please specify)
        \end{itemize}

        \\

        Q17 & \textbf{What was the type of the summary?}
        \begin{itemize}[label=$\square$, leftmargin=*, nosep]
          \item Lecture notes
          \item Blog post
          \item Highlights of some kind in the original study material
          \item Abstractive piece of text, such as a written overview of a text book, an abstract of a paper, etc.
          \item Short video
          \item A slide show
          \item Other (please specify)
        \end{itemize}

        \\

        Q18 & \textbf{How was the summary structured? (Multiple answers possible)}
        \begin{itemize}[label=$\square$, leftmargin=*, nosep]
          \item The summary was a running text, without particular structure
          \item The summary consisted of highlights in the original study material, without particular structure
          \item The summary itself contained special formatting, such as bold or cursive text, highlights, etc
          \item The summary contained diagrams
          \item The summary contained tables
          \item The summary contained graphs
          \item The summary contained figures
          \item The summary contained headings
          \item The summary consisted of different sections / paragraphs
          \item Other (please specify)
        \end{itemize}

        \\

        Q19 & \textbf{How much of the study material was covered by the summary?}
        \begin{tabularx}{0.7\columnwidth}{
          !{\hskip 2pt}>{\centering\arraybackslash}p{0.13\columnwidth}
          !{\hskip 2pt}>{\centering\arraybackslash}p{0.13\columnwidth}
          !{\hskip 2pt}>{\centering\arraybackslash}p{0.13\columnwidth}
          !{\hskip 2pt}>{\centering\arraybackslash}p{0.13\columnwidth}
          !{\hskip 2pt}>{\centering\arraybackslash}p{0.13\columnwidth}}

          None of the study material was covered &
          Almost none of the study material was covered &
          Some of the study material was covered &
          Most of the study material was covered &
          All of the study material was covered \\
          (1) & (2) & (3) & (4) & (5) \\

          $\square$ & $\square$ & $\square$ & $\square$ & $\square$
        \end{tabularx}

        \\

        Q20 & \textbf{What was the style of this summary?}
        \begin{itemize}[label=$\square$, leftmargin=*, nosep]
          \item Informative: the summary simply conveyed the information that was in the original study material
          \item Indicative: the summary gave an idea of the topic of the study material, but not more
          \item Critical: the summary gave a critical review of the study material
          \item Aggregative: the summary put different source texts in relation to one another and by doing this gave an overview of a certain topic
          \item Other (please specify)
        \end{itemize}

        \\

        Q21 & \textbf{Overall, how helpful was the pre-made summary for you? Your score can range from (1) Not helpful at all, to (5) Very helpful.}

        \begin{tabularx}{0.7\columnwidth}{
          !{\hskip 2pt}>{\centering\arraybackslash}p{0.13\columnwidth}
          !{\hskip 2pt}>{\centering\arraybackslash}p{0.13\columnwidth}
          !{\hskip 2pt}>{\centering\arraybackslash}p{0.13\columnwidth}
          !{\hskip 2pt}>{\centering\arraybackslash}p{0.13\columnwidth}
          !{\hskip 2pt}>{\centering\arraybackslash}p{0.13\columnwidth}}

          Not helpful at all & & & & Very helpful \\
          (1) & (2) & (3) & (4) & (5) \\

          $\square$ & $\square$ & $\square$ & $\square$ & $\square$
        \end{tabularx}

        \\

        Q22 & \textbf{Imagine you could turn this summary into your ideal summary. What would you change?} \\

        & Open response

        \\

        Q23 & \textbf{Is there anything else you want us to know about the summary that we have not covered yet?} \\

        & Open response

        \\

        Q24 & \textbf{Is there anything else you want us to know about the original study material that we have not covered yet?} \\

        & Open response \textit{-- participants are led to Q40}

        \\ \midrule

        \multicolumn{2}{l}{\textbf{Start branch of participants who described an imagined summary}} \\ \midrule

        & Please take one of these study activities in mind and imagine you would have had a pre-made summary. \\

        Q25 & \textbf{The original study material consisted of}
        \begin{itemize}[label=$\square$, leftmargin=*, nosep]
          \item Mainly text \textit{-- participants are led to Q27}
          \item Mainly figures \textit{-- participants are led to Q26}
          \item Mainly video \textit{-- participants are led to Q26}
          \item Mainly audio \textit{-- participants are led to Q26}
          \item A combination of some or all of the above \textit{-- participants are led to Q26}
          \item Other (please specify) \textit{-- participants are led to Q26}
        \end{itemize}

        \\

        Q26 & \textbf{For now we narrow down our survey to study material that is mostly textual. Do you recall any other recent study activity where you could have used a pre-made summary and where the original study material mainly consisted of text?}
        \begin{itemize}[label=$\square$, leftmargin=*, nosep]
          \item Yes \textit{-- participants are led to Q27}
          \item No \textit{-- participants are led to Q48}
        \end{itemize}

        \\

        Q27 & \textbf{What was the goal of this study activity?}
        \begin{itemize}[label=$\square$, leftmargin=*, nosep]
          \item Studying for an exam
          \item Writing a paper / essay / report / etc.
          \item Doing homework exercises
          \item Other (please specify)
        \end{itemize}

        \\

        & Now some questions will follow about what the study material that could be summarized looked like. \\

        Q28 & \textbf{What was the length of the study material?}
        \begin{itemize}[label=$\square$, leftmargin=*, nosep]
          \item A single article
          \item Multiple articles
          \item A single book chapter
          \item Multiple book chapters from the same book
          \item Multiple book chapters from various books
          \item A combination of the above
          \item Other (please specify)
        \end{itemize}

        \\

        Q29 & \textbf{How was the study material structured? (Multiple answers possible)}
        \begin{itemize}[label=$\square$, leftmargin=*, nosep]
          \item There was no particular structure - e.g. just one large text
          \item The text contained a title or titles
          \item The text contained subheadings
          \item The text consisted of different chapters
          \item The text consisted of different sections and / or paragraphs
          \item Other (please specify)
        \end{itemize}

        \\

        Q30 & \textbf{What was the genre of the study material?}
        \begin{itemize}[label=$\square$, leftmargin=*, nosep]
          \item Mainly educational (such as a text book (chapter))
          \item Mainly scientific (such as an academic article, publication, report, etc)
          \item Mainly nonfiction writing (such as (auto)biographies, history books, etc)
          \item Mainly fiction writing (such as novels, short fictional stories, etc)
          \item Other (please specify)
        \end{itemize}

        \\

        Q31 & \textbf{How would you classify the difficulty level of the study material?}
        \begin{itemize}[label=$\square$, leftmargin=*, nosep]
          \item Ordinary: most people would be able to understand it
          \item Specialized: you need to know the jargon of the field to be able to understand it
          \item Geographically based: you can only understand it if you are familiar with a certain area, for example because it contains local names
        \end{itemize}

        \\

        & Now we will ask some questions about the purpose of the pre-made summary that would have been helpful. \\

        Q32 & \textbf{For what type of people should the summary ideally be intended? Your score can range from (1) Untargetted, to (5) Targetted.}
        \begin{tabularx}{0.7\columnwidth}{
          !{\hskip 2pt}>{\centering\arraybackslash}p{0.13\columnwidth}
          !{\hskip 2pt}>{\centering\arraybackslash}p{0.13\columnwidth}
          !{\hskip 2pt}>{\centering\arraybackslash}p{0.13\columnwidth}
          !{\hskip 2pt}>{\centering\arraybackslash}p{0.13\columnwidth}
          !{\hskip 2pt}>{\centering\arraybackslash}p{0.13\columnwidth}}

          Untargetted: No domain knowledge is expected from the users of the summmary. & & & & Targetted: Full domain knowledge is expected from the users of the summmary. \\
          (1) & (2) & (3) & (4) & (5) \\

          $\square$ & $\square$ & $\square$ & $\square$ & $\square$
        \end{tabularx}

        \\

        Q33 & \textbf{How would this summary help you with your task? (Multiple answers possible)}
        \begin{itemize}[label=$\square$, leftmargin=*, nosep]
          \item The summary would help to retrieve parts of the original study material
          \item I would use the summary to preview the text that I was about to read
          \item I would use the summary as a substitute for the original study material
          \item I would use the summary to refresh my memory of the original study material
          \item I would use the summary as a reminder that I had to read the original study material
          \item The summary would help to get an overview of the original study material
          \item The summary would help to understand the original study material',
          \item Other (please specify)
        \end{itemize}

        \\

        & Now we will ask some questions about what the summary should look like and cover. \\

        Q34 & \textbf{What would be the ideal type of the summary?}
        \begin{itemize}[label=$\square$, leftmargin=*, nosep]
          \item Lecture notes
          \item Blog post
          \item Highlights of some kind in the original study material
          \item Abstractive piece of text, such as a written overview of a text book, an abstract of a paper, etc.
          \item Short video
          \item A slide show
          \item Other (please specify)
        \end{itemize}

        \\

        Q35 & \textbf{What is the ideal structure of the summary? (Multiple answers possible)}
        \begin{itemize}[label=$\square$, leftmargin=*, nosep]
          \item The summary should be a running text, without particular structure
          \item The summary should consist of highlights in the original study material, without particular structure
          \item The summary itself should contain special formatting, such as bold or cursive text, highlights, etc.
          \item The summary should contain diagrams
          \item The summary should contain tables
          \item The summary should contain graphs
          \item The summary should contain figures
          \item The summary should contain headings
          \item The summary should consist of different sections / paragraphs
          \item Other (please specify)
        \end{itemize}

        \\

        Q36 & \textbf{How much of the study material should be covered by the summary?}
        \begin{tabularx}{0.7\columnwidth}{
          !{\hskip 2pt}>{\centering\arraybackslash}p{0.13\columnwidth}
          !{\hskip 2pt}>{\centering\arraybackslash}p{0.13\columnwidth}
          !{\hskip 2pt}>{\centering\arraybackslash}p{0.13\columnwidth}
          !{\hskip 2pt}>{\centering\arraybackslash}p{0.13\columnwidth}
          !{\hskip 2pt}>{\centering\arraybackslash}p{0.13\columnwidth}}

          None of the study material should be covered &
          Almost none of the study material should be covered &
          Some of the study material should be covered &
          Most of the study material should be covered &
          All of the study material should be covered \\
          (1) & (2) & (3) & (4) & (5) \\
          $\square$ & $\square$ & $\square$ & $\square$ & $\square$
        \end{tabularx}

        \\

        Q37 & \textbf{What should the style of this summary be?}
        \begin{itemize}[label=$\square$, leftmargin=*, nosep]
          \item Informative: the summary should simply convey the information that was in the original study material
          \item Indicative: the summary should give an idea of the topic of the study material, but not more
          \item Critical: the summary should give a critical review of the study material
          \item Aggregative: the summary should put different source texts in relation to one another and by doing this give an overview of a certain topic
          \item Other (please specify)
        \end{itemize}

        \\

        Q38 & \textbf{Is there anything else you would want us to know about your ideal summary that we have not covered yet?} \\

        & Open response

        \\

        Q39 & \textbf{Is there anything else you would want us to know about the original study material that we have not covered yet?} \\

        & Open response

        \\ \midrule

        \multicolumn{2}{l}{\textbf{Look out questions}} \\ \midrule

        & Now, let's assume the pre-made summary was generated by a computer. You can assume that this machine generated summary captures all the needs you have identified in the previous questions. \\

        Q40 & \textbf{Would it make a difference to you whether the summary was generated by a computer program or by a human?}
        \begin{itemize}[label=$\square$, leftmargin=*, nosep]
          \item Yes \textit{-- participants are led to Q41}
          \item No \textit{-- participants are led to Q43}
        \end{itemize}

        \\

        Q41 & \textbf{Please explain the difference.} \\

        & Open response

        \\

        Q42 & \textbf{Which type of summary would you trust more:}
        \begin{itemize}[label=$\square$, leftmargin=*, nosep]
          \item A summary generated by a human, for example a teacher or a good performing fellow student
          \item A summary generated by a computer
          \item No difference
        \end{itemize}
        \\

        Q43 & \textbf{Please explain your answer.} \\

        & Open response

        \\

        Q44 & \textbf{Which type of summary would you trust more:}
        \begin{itemize}[label=$\square$, leftmargin=*, nosep]
          \item A summary generated by a human, for example a teacher or a good performing fellow student
          \item A summary generated by a computer
          \item No difference
        \end{itemize}

        \\

        & Now imagine that you can interact with the computer program that made the summary, in the form of a digital assistant. Imagine that your digital assistant made an initial summary for you and you can ask questions about it to your digital assistant and the assistant can answer them. Answers can be voice output, but also screen output, e.g. a written summary on the screen. In the next part we would like to investigate how you would interact with the assistant. Please do not feel restricted by the capabilities of today's digital assistants. \\

        Q45 & \textbf{Please choose the three most useful features for a digital assistant to have in this scenario.}
        \begin{itemize}[label=$\square$, leftmargin=*, nosep]
          \item Summarize particular parts of the study material with more detail
          \item Summarize particular parts of the study material with less detail
          \item Switch between different summary styles (for example highlighting vs a generated small piece of text)
          \item Explain why particular pieces ended up in the summary
          \item Provide the source of certain parts of the summary on request
          \item Search for different related sources based on the content of the summary
          \item Answer specific questions based on the content of the summary
        \end{itemize}

        \\

        Q46 & \textbf{Please choose the three least useful features for a digital assistant to have in this scenario.}
        \begin{itemize}[label=$\square$, leftmargin=*, nosep]
          \item Summarize particular parts of the study material with more detail
          \item Summarize particular parts of the study material with less detail
          \item Switch between different summary styles (for example highlighting vs a generated small piece of text)
          \item Explain why particular pieces ended up in the summary
          \item Provide the source of certain parts of the summary on request
          \item Search for different related sources based on the content of the summary
          \item Answer specific questions based on the content of the summary
        \end{itemize}

        \\

        Q47 & \textbf{Can you think of any other features that you would like your digital assistant to have to help you in this scenario?} \\

        & Open response

        \\ \midrule

        \multicolumn{2}{l}{\textbf{Background questions}} \\ \midrule

        & Thank you for filling out this survey so far! We would still like to ask you two final background questions. \\

        Q48 & \textbf{What is the current level of education you are pursuing?}
        \begin{itemize}[label=$\square$, leftmargin=*, nosep]
          \item Bachelor's degree
          \item Master's degree
          \item MBA
          \item Other, please specify
        \end{itemize}

        \\

        Q49 & \textbf{What is your main field of study?} \\

        & Open response \\

        \\ \midrule

        \multicolumn{2}{l}{\textbf{Thank you!}} \\ \midrule

        & You have come to the end of our survey. Thanks a lot for helping out! We very much appreciate your time. \\

        Q50 & \textbf{If you would like to participate in the raffle to win a voucher, please fill out your e-mail address below. We will only use this e-mail address to blindly draw 10 names who win a voucher and to contact you if your name has been drawn.} \\

        & Open response

\end{xtabular}

\twocolumn


\section{Full results trustworthiness and future feature questions}
\label{sec:appendix_future_features}

\parindent 1em

\begin{figure*}
  \centering
   \begin{subfigure}[b]{0.31\textwidth}
       \centering
       \caption{Would it make a difference to you whether the summary was generated by a computer program or by a human? (MC)}
       \vspace{-2pt}
       \includegraphics[clip,trim=0mm 0mm 0mm 10mm,width=\textwidth]{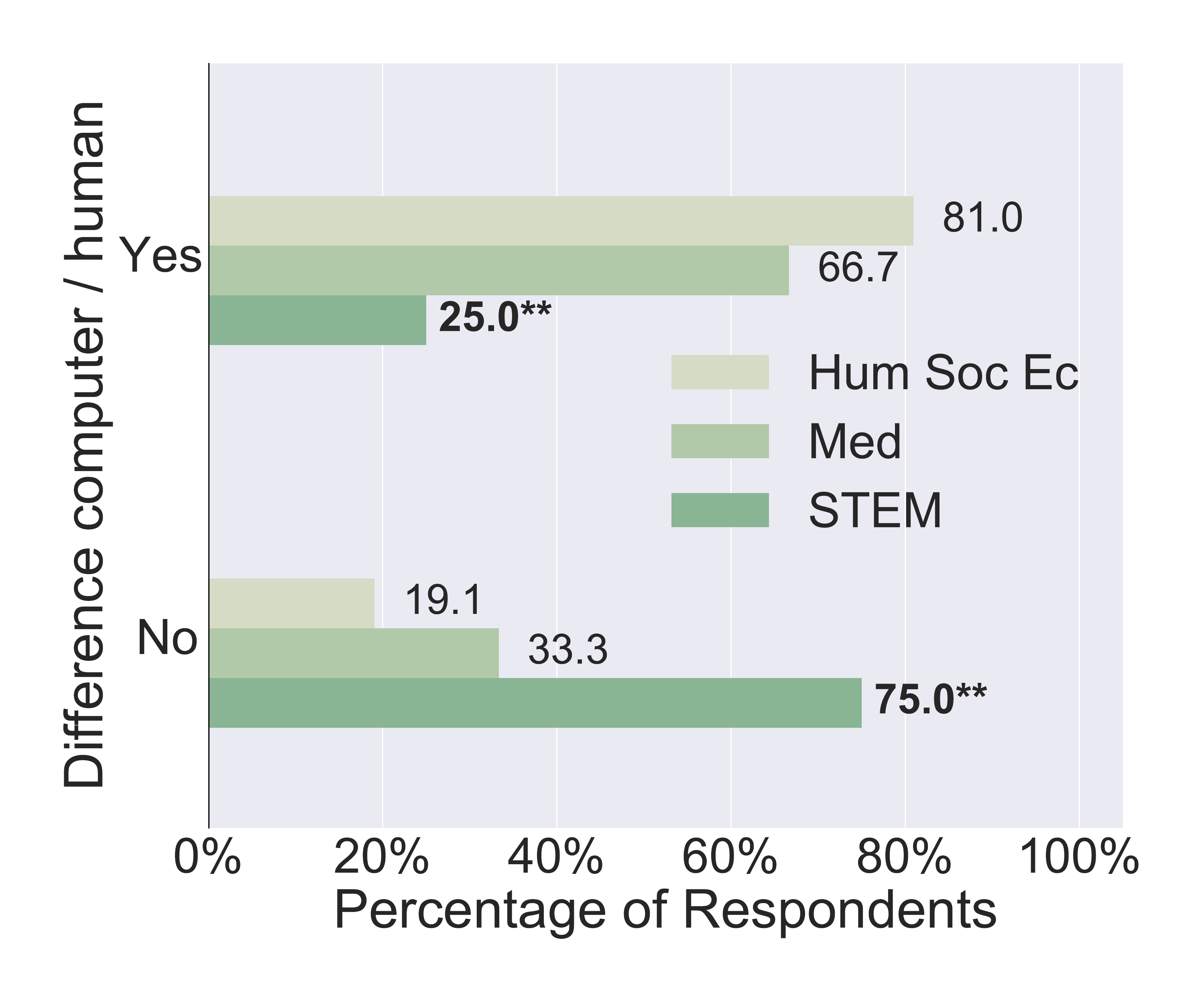}
       \label{fig:difference_human_computer}
       \vspace*{-0.5cm}
   \end{subfigure}\quad
   \begin{subfigure}[b]{0.31\textwidth}
       \centering
       \caption{Which type of summary would you trust more?~(MC) \phantom{two} \phantom{three} \phantom{four} \phantom{a} \phantom{bunch} \phantom{of} \phantom{random words}\\}
       \vspace{-2pt}
       \includegraphics[clip,trim=0mm 0mm 0mm 10mm,width=\textwidth]{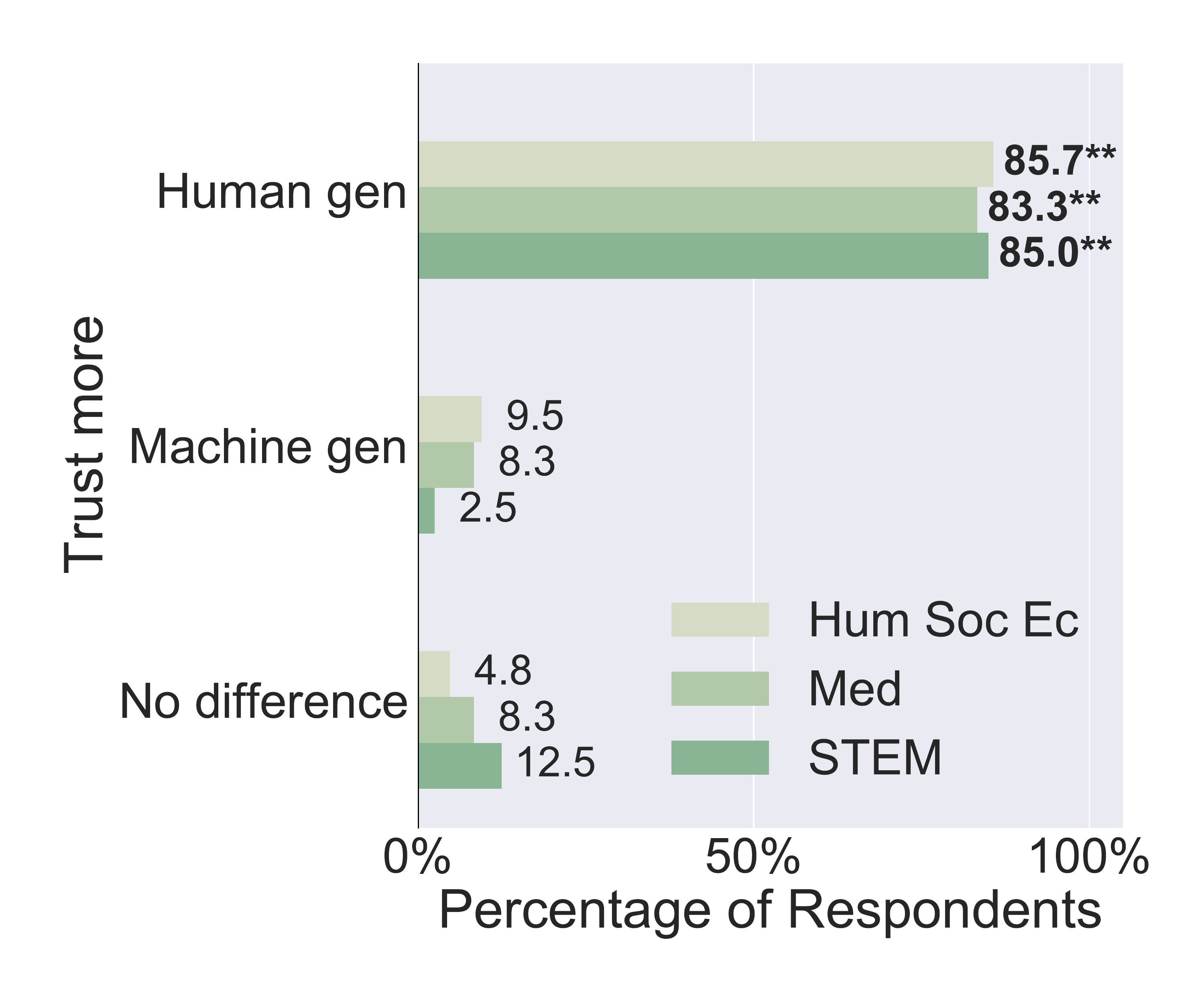}
       \label{fig:trust_more}
       \vspace*{-0.5cm}
   \end{subfigure}\quad

   \vspace{-1em}

   \begin{subfigure}[b]{0.31\textwidth}
       \centering
       \caption{Please choose the three \textit{most} useful features for a digital assistant to have in this scenario. (MR)}
       \vspace{-2pt}
       \includegraphics[clip,trim=0mm 0mm 0mm 10mm,width=\textwidth]{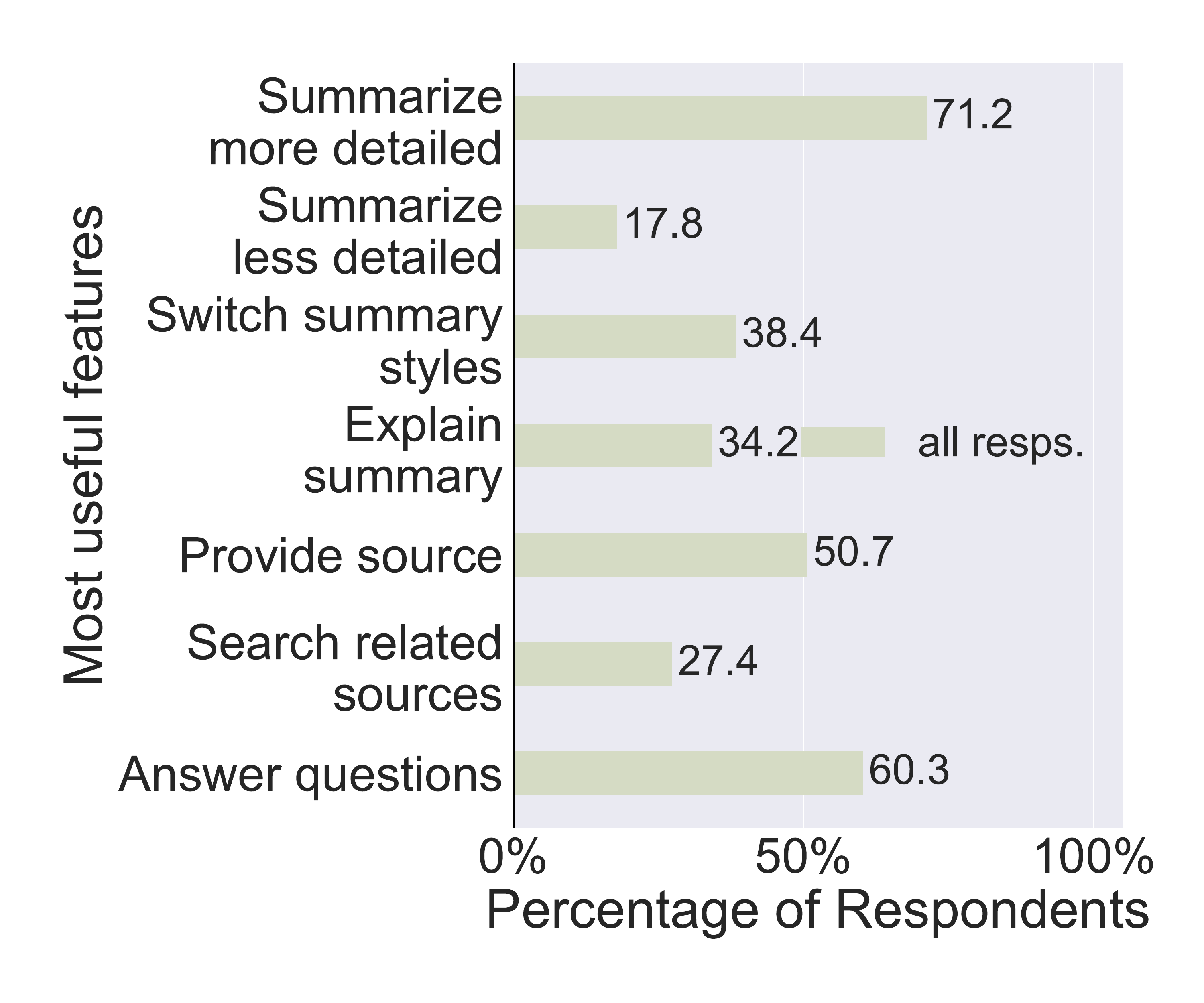}
       \label{fig:most_useful_features}
       \vspace*{-0.5cm}
   \end{subfigure}\quad
   \begin{subfigure}[b]{0.31\textwidth}
       \centering
       \caption{Please choose the three \textit{least} useful features for a digital assistant to have in this scenario. (MR)}
       \vspace{-2pt}
       \includegraphics[clip,trim=0mm 0mm 0mm 10mm,width=\textwidth]{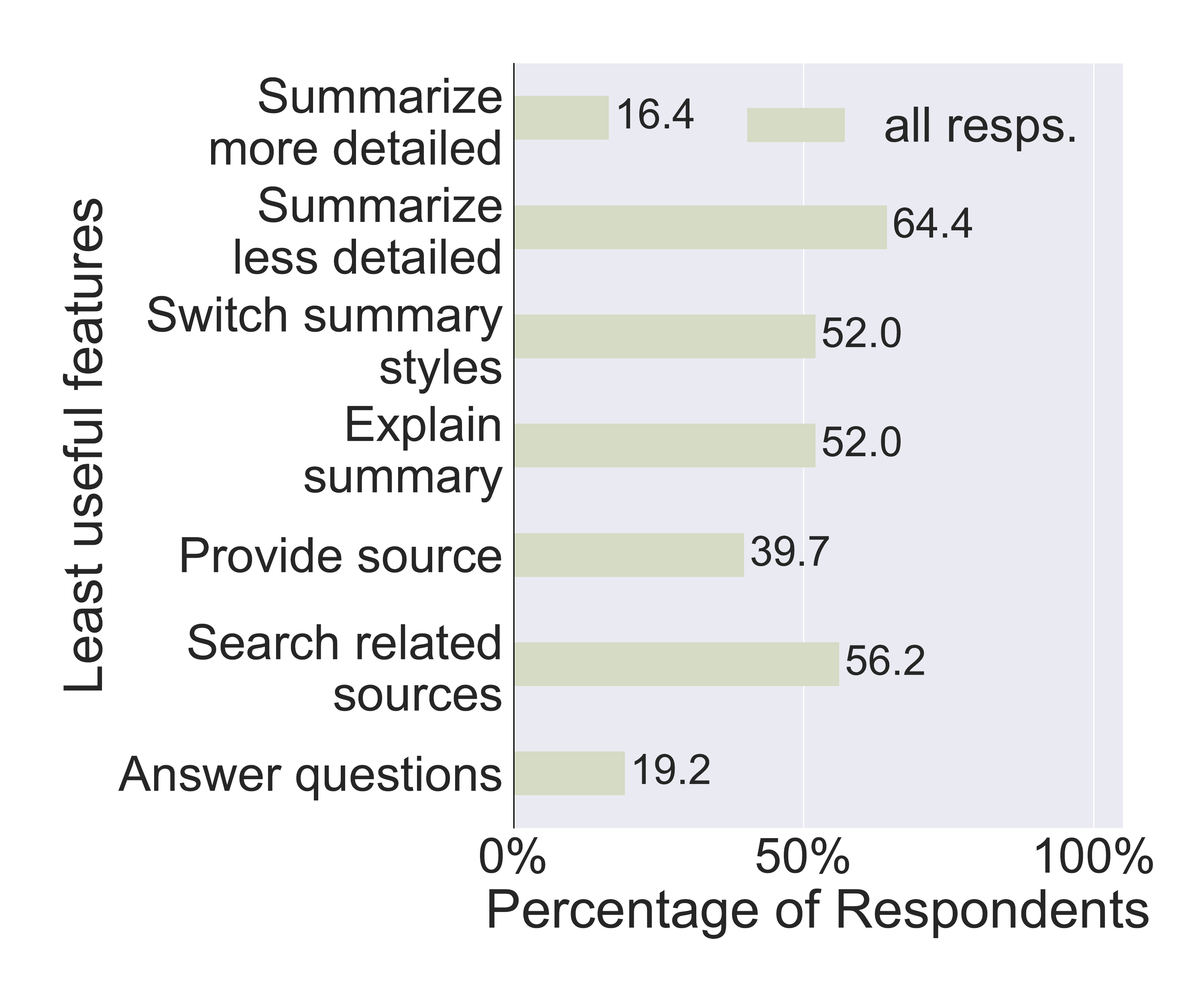}
       \label{fig:least_useful_features}
       \vspace*{-0.5cm}
    \end{subfigure}\quad

\caption{Results for the future feature questions. Answer type in brackets. MC = Multiple Choice, MR = Multiple Response. \textbf{**} indicates significance ($\chi^2$ or Fisher's exact test), after Bonferroni correction, with $p\ll 0.001$.}
\label{fig:lookout_questions}
\vspace{-5pt}
\end{figure*}

In this section we report the results for the exploratory questions that we asked about the trustworthiness of a summary generated by a machine versus a human, as well as the results for the questions about features for summarization with a digital voice assistant.

We find that participants are divided on the question whether it would make a difference to them whether the summary was generated by a machine or a computer.
If we look at all participants together, we find that $48.0.\%$ of the participants answered that it would make a difference, whereas $52.0\%$ answered that it would not. However, if we split the participants based on study background, an interesting difference emerges (Figure~\ref{fig:difference_human_computer}). Participants with a background in STEM indicated significantly more often that it would not make a difference to them, whereas the other groups of students indicated the opposite. Almost all participants who answered that it would make a difference said that they would not trust a computer on being able to find the relevant information, i.e., all seemed to favor the human generated summary. Only one participant advocated for the computer-generated summary as a \textit{``computer is more objective.''} Almost all participants who said it would not matter to them did add the condition that the quality of the generated summary should be as good as if a human had generated it.
One person wrote: \textit{``If the summary captures all previously discussed elements it is effectively good for the same purpose. So then it does not matter who generated it.''} This comment exactly captures the motivation of the setup of our survey.

This caution regarding automatically generated summaries is confirmed by the question in which we asked which type of summary participants would trust more -- a human-generated one or a machine-generated one. People chose the human-generated summary significantly more often (Figure~\ref{fig:trust_more}). This also holds for the participants with a STEM background, which aligns with the responses to the open questions we reported earlier -- apparently participants do not fully trust that the condition they raised earlier would be satisfied, namely that only if the machine was just as good as the human, it would not matter for them whether the summary was generated by a machine or a human.

The results for the most and least useful features for a digital assistant in a summarization scenario are given in Figure~\ref{fig:most_useful_features} and ~\ref{fig:least_useful_features}. Adding more details to the summary and answering questions based on the content of the summary are very popular features, whereas summarizing parts of the input material with less detail is not.

Lastly, we asked participants whether they could think of any other features that they would like their digital assistant to have in the outlined scenario. A number of participants answered that they would like the digital assistant to generate questions based on the summary, so that they could test their own understanding. For example, one participant said: \textit{``Make questions for me (to test me)''} and another participant had a related comment: \textit{``Maybe the the digital assistant could find old exam questions to link to parts of the summary where the question is related to, so that there is a function to test if you've understood the summary.''} Another line of answers pointed towards giving explicit relations between the input material and summary, for example: \textit{``Show links between subject materials and what their relation is''} and another person wrote: \textit{``Dynamic linking from summary to original source is a great added value of generating a summary''}.


\section{Examples evaluation questions}
\label{sec:appendix_evaluation_questions}

Here we give additional examples for the evaluation questions that can be used for our proposed evaluation methodology. The phrase \textit{``a document that is important for your task"} should be substituted to match the task at hand. For example, in the case of exam preparations, this could be replaced with \textit{``a chapter that you need to learn for your exam preparation"}. Only the questions with the intended purpose factors should be used in the evaluation.
\bigskip

\textbf{Purpose factor \textit{Use} \& Output factor \textit{Style}}:
\begin{itemize}[leftmargin=*, nosep]
  \item The \textit{style} of which of these two summaries is most useful to you to \textit{retrieve} a document that is important for your task?
  \item The \textit{style} of which of these two summaries is most useful to you to \textit{preview} a document that is important for your task?
  \item The \textit{style} of which of these two summaries is most useful to you to \textit{substitute} a document that is important for your task?
  \item The \textit{style} of which of these two summaries is most useful to you to \textit{refresh your memory} about a document that is important for your task?
  \item The \textit{style} of which of these two summaries is most useful to you to \textit{prompt} you to read a source text that is important for your task?
\end{itemize}
\bigskip
\noindent \textbf{Purpose factor \textit{Use} \& Output factor \textit{Format}}:
\begin{itemize}[leftmargin=*, nosep]
  \item The \textit{format} of which of these two summaries is most useful to you to \textit{retrieve} a document that is important for your task?
  \item The \textit{format} of which of these two summaries is most useful to you to \textit{preview} a document that is important for your task?
  \item The \textit{format} of which of these two summaries is most useful to you to \textit{substitute} a document that is important for your task?
  \item The \textit{format} of which of these two summaries is most useful to you to \textit{refresh your memory} about a document that is important for your task?
  \item The \textit{format} of which of these two summaries is most useful to you to \textit{prompt} you to read a source text that is important for your task?
\end{itemize}
\bigskip
\noindent \textbf{Purpose factor \textit{Use} \& Output factor \textit{Material}}:
\begin{itemize}[leftmargin=*, nosep]
  \item The \textit{coverage} of which of these two summaries is most useful to you to \textit{retrieve} a document that is important for your task?
  \item The \textit{coverage} of which of these two summaries is most useful to you to \textit{preview} a document that is important for your task?
  \item The \textit{coverage} of which of these two summaries is most useful to you to \textit{substitute} a document that is important for your task?
  \item The \textit{coverage} of which of these two summaries is most useful to you to \textit{refresh your memory} about a document that is important for your task?
  \item The \textit{coverage} of which of these two summaries is most useful to you to \textit{prompt} you to read a source text that is important for your task?
\end{itemize}

\end{document}